\documentclass[10pt,twocolumn,letterpaper]{article}

\usepackage{cvpr}
\usepackage{times}
\usepackage{epsfig}
\usepackage{graphicx}
\usepackage{amsmath}
\usepackage{amssymb}
\usepackage{threeparttable}
\usepackage{mathrsfs}
\usepackage[ruled]{algorithm2e}
\usepackage{subfigure}
\usepackage{multirow}

\usepackage[breaklinks=true,bookmarks=false]{hyperref}

\cvprfinalcopy 

\def\cvprPaperID{****} 

\begin{document}

\title{Histopathology WSI Encoding based on GCNs for Scalable and Efficient Retrieval of Diagnostically Relevant Regions}

\author{Yushan Zheng, Zhiguo Jiang, Haopeng Zhang, Fengying Xie\\
 Beijing Advanced Innovation Center for Biomedical Engineering, Beihang University, China\\
{\tt\small yszheng@buaa.edu.cn jiangzg@buaa.edu.cn                                                                                                                                           }
\and
Jun Shi\\
School of Software, Hefei University of Technology, China\\
\and
Chenghai Xue\\
Tianjin Institute of Industrial Biotechnology, Chinese Academy of Sciences, China\\
}

\maketitle

\begin{abstract}
   	Content-based histopathological image retrieval (CBHIR) has become popular in recent years in the domain of histopathological image analysis. CBHIR systems provide auxiliary diagnosis information for pathologists by searching for and returning regions that are contently similar to the region of interest (ROI) from a pre-established database. While, it is challenging and yet significant in clinical applications to retrieve diagnostically relevant regions from a database that consists of histopathological whole slide images (WSIs) for a query ROI. In this paper, we propose a novel framework for regions retrieval from WSI-database based on hierarchical graph convolutional networks (GCNs) and Hash technique. Compared to the present CBHIR framework, the structural information of WSI is preserved through graph embedding of GCNs, which makes the retrieval framework more sensitive to regions that are similar in tissue distribution. Moreover, benefited from the hierarchical GCN structures, the proposed framework has good scalability for both the size and shape variation of ROIs. It allows the pathologist defining query regions using free curves according to the appearance of tissue. Thirdly, the retrieval is achieved based on Hash technique, which ensures the framework is efficient and thereby adequate for practical large-scale WSI-database. The proposed method was validated on two public datasets for histopathological WSI analysis and compared to the state-of-the-art methods. The proposed method achieved mean average precision above 0.857 on the ACDC-LungHP dataset and above 0.864 on the Camelyon16 dataset in the irregular region retrieval tasks, which are superior to the state-of-the-art methods. The average retrieval time from a database within 120 WSIs is 0.802 ms.
\end{abstract}

\section{Introduction}
With the development of digitalization techniques for pathology, the computer-aided cancer diagnosis methods based on histopathlogical image analysis (HIA) \cite{litjens2017survey,gurcan2009histopathological} have been widely studied. In recent years, the researches are generally focused on the histopathological image classification~\cite{zheng2017feature,Xu2017Large}, segmentation \cite{Xu2014Weakly,bejnordi2016automated,bejnordi2017diagnositic,jia2017constrained,falk2019u} and object detection \cite{xu2015stacked, veta2019predicting}. The applications can provide the pathologists a definitely suggestion for diagnosis and even automatically generate a quantitative report for the input case. The output is flat and visualized. 
It determines the models are convenient to deployment in digital pathology platforms. Nevertheless, the limitation of these methods is also apparent. Generally, models of image classification, segmentation and detection can hardly provide additional diagnosis information beyond the flat model output. Specifically, these methods predict where and what the cancer it is but can hardly provide the dependence or reason of the decision.

Content-based histopathological image retrieval (CBHIR) is an emerging approach in the domain of HIA~\cite{zhang2016large-scale, li2018large-scale}. CBHIR derived from content-based image retrieval (CBIR), which is another popular methodology besides image classification, segmentation and detection in natural scene image processing. CBHIR searches for a pre-established WSI database for the regions the pathologist concerned and provides contently similar regions to the pathologists for reference. Compared to the typical HIA methods mentioned above, CBHIR methods provide more valuable information including similar regions from diagnosed cancer cases, the corresponding meta-information and the diagnosis reports of experts stored along with the cases in the digital pathology platform. The informative output can be regarded as the dependence of the automatic diagnosis, which is of developmental significance to pathologists. 

In recent five years, the techniques of whole slide imaging are well developed and gradually popularized in clinical diagnosis. The size of histopathological whole slide images (WSIs) database are rapidly increased. In this situation, it is crucial and yet challenging to develop effective methods for retrieving diagnostically relevant sub-regions from large-scale WSI-database. 

In this paper, we propose a novel CBHIR framework for diagnostically relevant region retrieval from large-scale WSI-database based on graph convolutional network (GCN)~\cite{wu2019comprehensive} and hashing method. Different from the present sub-regions retrieval frameworks~\cite{jim2017deep,ma2018generating,zheng2018histopathological,zheng2018size-scalable}, we propose constructing graphs for the sub-regions in the WSI to describe the structural information within the regions and employing GCNs with differentiable pooling modules to encode the graph features into uniform retrieval indexes. Both the local features and structural information in the regions are effectively preserved in the encoding. Moreover, we modified the GCN model with hashing methodology to ensure the efficiency of retrieval. The proposed method was evaluated on two public available histopathological WSI datasets and compared with the state-of-the-art methods developed for histoapthological image retrieval. The experimental results have demonstrated the effectiveness and efficiency of our method.

The contribution of this paper to the problem is three-fold:

1) We proposed a novel WSI encoding methods based on graph convolutional networks. To our knowledge, we are the first to use GCNs for histopathological image retrieval. Besides the local features of tissue regions, the distribution of tissue is considered in the process of WSI encoding. The accuracy and scalablility of CBHIR has been significantly improved. 

2) We designed a GCN-Hash model by combining the GCN structure with the Hash model. The GCN-Hash model can be trained end-to-end from graphs with variable number of nodes to the binary-like codes and the retrieval can be achieved based on hamming distances between binary codes. The speed of retrieval has been effectively improved. It determines the proposed method is applicable to practical large-scale WSI database. 

3) We designed experiments to verify the proposed retrieval framework with CBIR metrics and compared it with 4 state-of-the-art retrieval framework proposed for histopathological images on two public accessible datasets involving two typical types of cancer. The experimental results have demonstrated the proposed GCN-Hash model have achieved the state-of-the-art retrieval performance.

The remainder of this paper is organized as follows. Section \ref{S.related} reviews the history of histopathological image retrieval. Section \ref{S.method} introduces the methodology of the proposed method. The experiment and discussion are presented in Section \ref{S.experiment}. Section \ref{S.conclusion} summarizes the contributions. This is an expanded version of the conference paper \cite{zheng2019encoding}.

\begin{figure*} [t]
	\includegraphics[width=\textwidth]{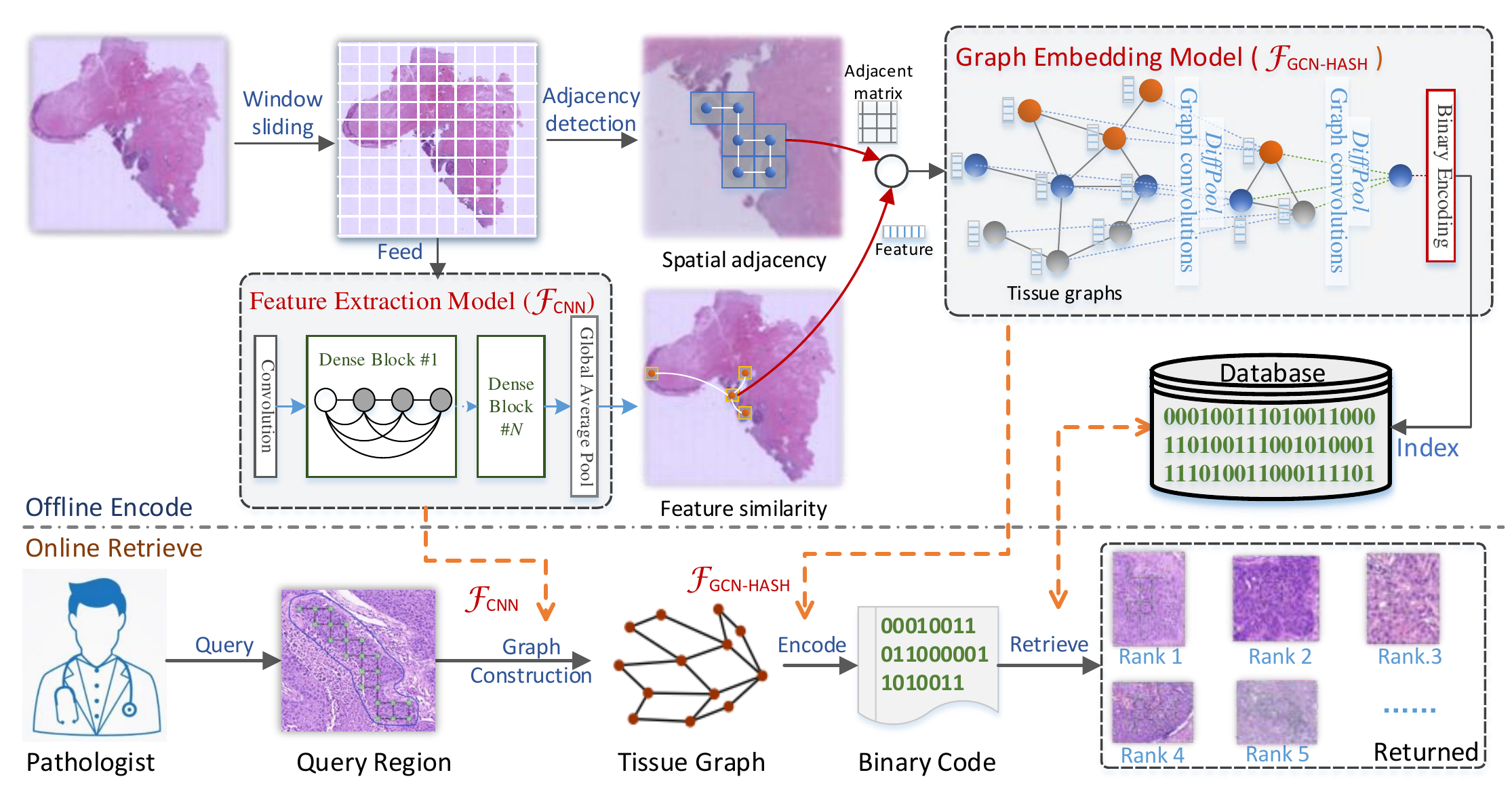}
	\caption{The proposed CBHIR framework. In the offline stage, the WSI is first divided into patches following the sliding window paradigm and a CNN is trained based on the patches to extract image features. Then, tissue graphs are constructed based on the spatial adjacency and feature similarity of patches. Finally, GCNs are established to encode the graphs into binary codes, which are used to index the retrieval database. In the online retrieval stage, the region the pathologist queried are converted into a binary code using the trained models. The most relevant regions are retrieved by measuring the similarities between the query code and those in the database and finally returned to pathologists for diagnosis reference.} \label{F.framework}
\end{figure*}

\section{Related Works} \label{S.related}
The objects in the studies on CBHIR have been through cells/nuclei, image patches and whole slide images along with the development of digital pathology. The typical methods related to our work are reviewed in this section.

\subsection{Retrieval methods for cells and patches}
The early studies focused on the cell retrieval from histological images that were captured under the optical microscopy \cite{Comaniciu1998Bimodal,Comaniciu1998Shape,Wetzel1999Evaluation}.
With the development of the digitalization of histological sections, the CBHIR frameworks for patch-level retrieval were proposed. For example, The methods \cite{Zheng2004Design,Zhou2004Content,Mehta2009Content} employed the classical image features to depict the histopathological images and achieved the patch-level retrieval. Then, the retrieval methodology was studied in various aspects.

A number of works concentrated on extracting high-level features of histopathological images to improve the accuracy of retrieval. Specifically, CBHIR frameworks based on manifold learning \cite{Doyle2007Using,Sparks2011Out}, semantic analysis \cite{Caicedo2008A,Caicedo2010Combining,Zheng2014Retrieval}, spectral embedding \cite{Sridhar2011Boosted} and fine-designed local descriptors \cite{tizhoosh2018representing,erfankhah2019heterogeneity} have been developed and have proven effective in improving the accuracy of retrieval. Meanwhile, other works \cite{gu2018densely,zheng2018histopathological,gu2019multi} proposed utilizing the contextual information by combining features from multiple magnifications of histopathological images to enhance the representations of image patches and thus improve the performance of retrieval. As for the online usage of CBHIR, the security of retrieval has already been considered \cite{cheng2019histopathological}.

Besides the retrieval accuracy, the efficiency of CBHIR has become increasingly popular in the recent years. To satisfy the application for database consisting of massive histopathological images, hashing techniques were introduced. Typically, Zhang \etal \cite{Zhang2015Towards,zhang2015fusing,Jiang2016Scalable} introduced supervised hashing with kernels (KSH) \cite{Liu2012Supervised} into the CBHIR. Shi \etal \cite{Shi2017Supervised} utilized a graph hashing model to learn the similarity relationship of hitopathological images. With hashing functions, the images are encoded into an array of binary codes. And the similarities among images are measured by Hamming distance, which is able to be calculated very efficiently using bitwise operations by computer. More recently, The end-to-end deep learning frameworks \cite{shi2018pairwise,sapkota2018deep,peng2019multi} were constructed based on CNNs to directly encode histopathological images into binary codes. The overall performance of patch-level CBHIR has been further improved.

\subsection{Whole slide image database retrieval}
With the widely application of digital pathology, the present database in clinical applications usually consists of massive WSIs. It has become crucial to retrieve and return relevant sub-regions from the WSI-database for a region the pathologist provided during the diagnosis. 

In the previous study, Ma \etal \cite{ma2016breast} proposed dividing the WSIs into sub-regions following the sliding window paradigm and encoding the individual regions to establish the retrieval database. It is a convenient strategy to index WSIs for sub-regions retrieval. However, the tissue appearance was ignored in the division of WSIs and retrieval instances in the database were limited into rectangle images in fixed sizes. It gaps from the applicable situation where the ROIs are usually defined by pathologists with free-carves in various shapes and sizes. 

Then, several retrieval strategies have been developed to improve the scalability of the retrieval framework. Zheng \etal \cite{zheng2018histopathological} proposed segmenting a WSI into super-pixels and defining the super-pixels as retrieval instances. Further, Ma \etal \cite{ma2018generating} proposed merging the super-pixels into irregular regions based on selective search \cite{Uijlings2013Selective} to index the WSI. The query ROI in these methods was not restricted in rectangle regions. However, the representation of an irregular region was obtained by the max-pooling of the features of super-pixels. The size and shape of regions were not described. In the methods \cite{jim2017deep} and \cite{zheng2018size-scalable}, the ratio of histological objects has been considered by measuring the similarity between each pair of local features across two regions. Nevertheless, the adjacency relationship of different types of histological objects are hardly measured in these methods and thereby the structural similarity between tissue regions are difficult to be recognized in the retrieval procedure.

To conquest the drawbacks in the present methods, we proposed to construct graphs within the irregular sub-regions to depict the structure of tissue. Furthermore, motivated by the development of graph information analysis (e.g. molecules structure recognition and social network analysis) based on graph neural networks (GNNs) \cite{dai2016discriminative,kipf2017semi,gilmer2017neural,velivckovic2017graph,wu2019comprehensive}, we proposed to establish an end-to-end networks based on GNNs to encode the graphs into uniform indexes. Therefore, both the local features and adjacency relationships are hopefully preserved in the indexes and reflected in the results of retrieval. 
\section{Methods} \label{S.method}
\subsection{Overview}
The proposed CBHIR framework is illustrated in Fig.~\ref{F.framework}. The WSIs are first divided into patches and converted into image features using a pre-trained convolutional neural networks (CNN). Then, graphs of tissue are established based on spatial relationships and feature distances of patches. Finally, the tissue-graphs are fed into the designed GCN-Hash model to obtain the binary indexes for retrieval. In this section, the method for tissue graph construction is firstly introduced and then the GCN-Hash model will be presented.

\subsection{Tissue graph construction} \label{S.graph_construction}
\begin{figure*} [t]
	\includegraphics[width=\textwidth]{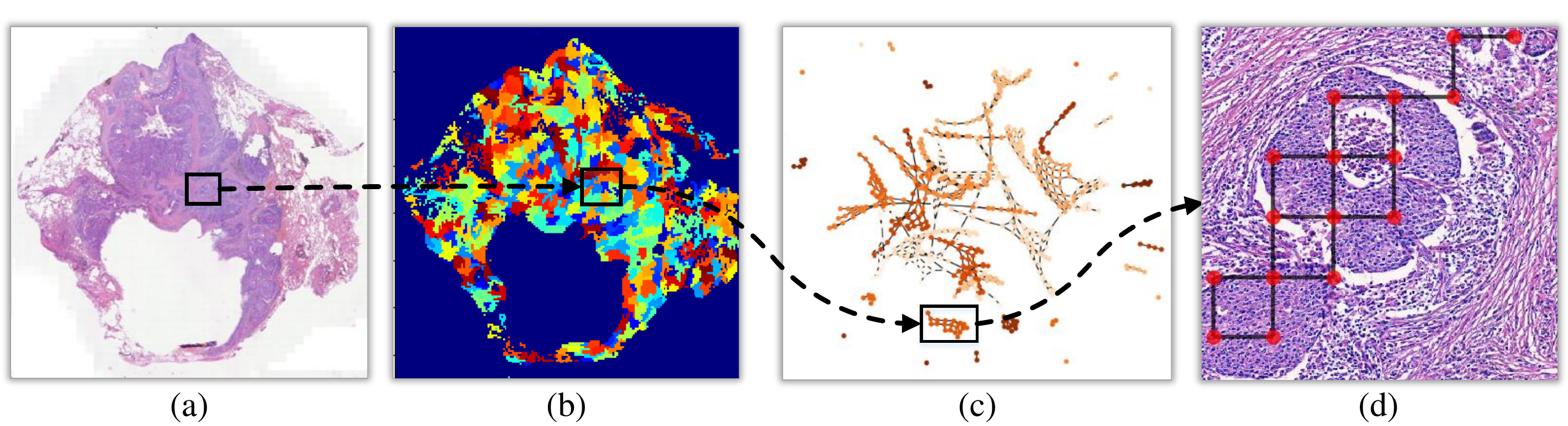}
	\caption{The flowchart of tissue graphs, where (a) is a digital WSI, (b) illustrates the sub-regions clustered by Algorithm \ref{A.tgc}, (c) shows the graphs established on the sub-regions, and (d) jointly presents a graph and its corresponding region where the nodes are drawn on the centers of patches.} \label{F.graph_construction}
\end{figure*}

In our method, the sub-regions in the WSI are described by graphs. The flowchart to construct graphs for a WSI is presented in Fig.~\ref{F.graph_construction}. The patches are fed into a pre-trained convolutional neural network (CNN) to extract patch features. Then, graphs are defined based on the sub-regions by regarding the patch as the graph vertex and the spatial adjacency as the graph edge. Therefore, a set of graphs are constructed (Fig.~\ref{F.graph_construction} (c)) for the WSI.

A graph $G$ is generally represented as $(\mathbf{A}, \mathbf{X})$, where $\mathbf{A}\in \mathbb{R}^{n\times n}$ is an adjacent matrix that defines the connectivity in $G$ with $n$ nodes, and $\mathbf{X}\in \mathbb{R}^{n\times d_f}$ denotes the node feature matrix assuming each node is represented as a $d_f$-dimensional vector.
For convenience, the graphs in the $s$-th WSI are represented by a set $\mathcal{G}_s=\{G_i|i=1,2,...,n_s\}$, where $n_s$ denotes the number of graphs in the $s$-th WSI. The set $\mathcal{G}_s$ can cover the entire content of the WSI and thereby are used to index the WSI for retrieval.  
There are two main techniques in stage of tissue graph construction. 

The one is the feature extractor for image patches based on CNN. CNN \cite{he2016deep, huang2017densely} has been well studied in recent years for which the methodology of CNN will not be detailed in this paper. Letting $I_i$ represent the $i$-th patch in a WSI, the process of feature extraction can be described as 
$$\mathbf{x}_i=\mathcal{F}_{CNN}(I_i),$$
where $\mathcal{F}_{CNN}$ represents a CNN feature extractor that takes a image patch as the input and outputs $d_f$-dimensional column vector.

The other main technique is the hierarchical agglomerative clustering (HAC) algorithm~\cite{Day1984Efficient}, which is employed in the flowchart to merge the patches and generate graphs. HAC is designed to merge a set of samples to an assigned number of clusters. In each iteration of HAC, the most similar two clusters under specific similarity measurement are combined. Specifically for WSIs, we propose regarding the feature $\mathbf{x}_i$ as the initial cluster and utilizing error sum of squares (EES)~\cite{ward1963hierarchical} as the similarity measurement between clusters. Besides, an adjacency matrix $\mathbf{A}_s\in\{0,1\}^{m_s\times m_s}$ is generated to indicate the connectivity of $m_s$ patches in the WSI, where $a_{ij}=1,a_{ij}\in\mathbf{A}_s$ indicates the $i$-th and the $j$-th patch are spatially 4-connected and $a_{ij}=0$ otherwise. To ensure the merged sub-regions are spatially connected, only the pairs associated with $a_{ij}=1$ are allowed to be merged in the iterations of HAC. The detailed algorithm of the tissue graph construction is defined in Algorithm \ref{A.tgc}.

\begin{algorithm} [t]\label{A.tgc}\small
	\caption{The algorithm of tissue graph construction.}
	\LinesNumbered
	\KwIn{\\
		$m_s\leftarrow$ The number of patches in the $s$-th WSI;\\
		$\{\mathbf{x}_i|i=1,2,...,m_s\}\leftarrow$ The feature vectors of patches;\\
		$\mathbf{A}_s\in\{0,1\}^{m_s\times m_s}\leftarrow$ The adjacency matrix of patches;\\
		$\hat{g}_s\leftarrow$ The target number of graphs ($\hat{g}_s \leq m_s$);}
	\KwOut{$\mathcal{G}_s$}
	\For{$i=1$ to $m_s$}
	{
		$C_i\leftarrow\{\mathbf{x}_{i}\}$; \\
	}
	$\mathcal{C}\leftarrow\{C_i|i=1,2,...,m_s\}$;\\
	$g_s\leftarrow m_s$;\\
	\While{$g_s>\hat{g}_s$}
	{
		$\mathcal{T}\leftarrow\o$;\\
		\For{$(C_i,C_j)$ in \\ $\quad\quad\{(C_i,C_j)|\exists \mathbf{x}_{p}\in C_i,\exists\mathbf{x}_{q}\in C_j, i\neq j, s.t. a_{pq}=1\}$}
		{
			$d_{ij}\leftarrow$\textit{EES}$(C_i \cup C_j)$; \\
			$\mathcal{T}\leftarrow \mathcal{T}\cup\{d_{ij}\}$;\\
		}
		index $(p,q)\leftarrow\arg\min_{(i,j)}(\mathcal{T})$;\\
		$C_p\leftarrow C_p\cup C_q$;\\
		$\mathcal{C}\leftarrow \mathcal{C}\setminus C_q$;\\
		$g_s \leftarrow g_s-1$;\\
	}
	$\mathcal{G}_s \leftarrow \o$;\\
	\For{$C_i$ in $\mathcal{C}$}
	{
		$\mathbf{X}_i\leftarrow(\mathbf{x}_1,...,\mathbf{x}_j,...,\mathbf{x}_{|C_i|})_{\mathbf{x}_j\in C_i}$;\\
		$\mathbf{A}_i\leftarrow$ Seek $\mathbf{A}_s$ for the adjacent relationship of patches corresponding to $\mathbf{X}_i$;\\
		$G_i \leftarrow (\mathbf{X}_i, \mathbf{A}_i)$;\\
		$\mathcal{G}_s \leftarrow \mathcal{G}_s \cup \{G_i\}$;\\
	}
	return $\mathcal{G}_s$;
\end{algorithm}

\subsection{GCN-Hash for region encoding} \label{S.gcn_hash}
It is challenging to simultaneously encode the node attributes and edge information into an uniform representation. 
In this paper, we propose a GCN-Hash model to encode sub-regions in WSIs. The algorithms involved in the GCN-Hash model will be introduced in this section.

\subsubsection{Graph convolutional network (GCN)}
GCN is in the scope of graph neural network (GNN). Generally, a GNN can be represented following message-passing architecture
\begin{equation}\label{E.graph_embed}
\mathbf{H}^{(k)}=M(\mathbf{A}, \mathbf{H}^{(k-1)}:\theta^{(k)}),
\end{equation}
where $\mathbf{H}^{(k)}\in \mathbb{R}^{n\times d_g}$ denotes the embedding on the $k$-th step of passing, $d_g$ denotes the dimension of the embedding, \textit{M} is the message propagation function~\cite{gilmer2017neural,hamilton2017inductive,kipf2017semi} that depends on the adjacent matrix $\mathbf{A}$, the output of the previous step $\mathbf{H}^{(k-1)}$ and the set of trainable parameters $\theta^{(k)}$. $\mathbf{H}^{(0)}$ is the original attributes of the nodes (i.e., the CNN features $\mathbf{X}$ in our method). In this paper, We apply the definition of graph convolutional networks \cite{kipf2017semi} to specializing the propagation function. Therefore, $M$ (Eq. \ref{E.graph_embed}) is formulated as 
\begin{equation}\label{E.gcn}
\mathbf{H}^{(k)}=ReLU(\tilde{\mathbf{D}}^{-\frac{1}{2}}\tilde{\mathbf{A}}\tilde{\mathbf{D}}^{-\frac{1}{2}}\mathbf{H}^{(k-1)}\mathbf{W}^{(k)}),
\end{equation}
where $\tilde{\mathbf{A}}=\mathbf{A}+\mathbf{E}$\footnote{$\mathbf{E}$ denotes the unit matrix.}, $\tilde{\mathbf{D}}=diag(\sum_j \tilde{\mathbf{A}}_{1j},\sum_j \tilde{\mathbf{A}}_{2j},...,\sum_j \tilde{\mathbf{A}}_{nj})$, and $\mathbf{W}^{(k)}\in\mathbb{R}^{m\times m}$ is a trainable weight matrix.

\subsubsection{Differentiable pooling for hierarchical GCNs}
Multiple GCNs can be stacked to learn hierarchical representations of graphs.
For simplicity, the $l$-th GCN module with $K$ embedding steps is represented as
$$\mathbf{Z}^{(l)}=\mathbf{H}^{(K)}=\mathcal{F}^{(l)}_{embed}(\mathbf{A}^{(l)}, \mathbf{X}^{(l)}),$$
where $\mathbf{A}^{(l)}$ and $\mathbf{X}^{(l)}$ denote the adjacency matrix and the input features of the $l$-th GCN, respectively, and $\mathbf{Z}^{(l)} \in \mathbb{R}^{n \times d}$ is the output of the GCN.

Recently, Ying \etal \cite{ying2018hierarchical} proposed a differentiable graph pooling module (DiffPool), which enables the hierarchical GCNs to be trained in end-to-end fashion. Specifically, an additional GCN with a row-softmax output layer is designed to learn an assignment matrix:
$$\mathbf{S}^{(l)}=softmax_r\left(\mathcal{F}^{(l)}_{pool}(\mathbf{A}^{(l)}, \mathbf{X}^{(l)})\right),$$
where $\mathcal{F}^{(l)}_{pool}$ represents a GCN under the same definition of $\mathcal{F}^{(l)}_{embed}$. $\mathbf{S}^{(l)}\in \mathbb{R}^{n_l\times n_{l+1}}$ ($n_{l+1}<n_l$) is used to assign the output representations of the $l$-th GCN to $n_{l+1}$ clusters. Then, the input $\mathbf{X}^{(l+1)}$ and the adjacent matrix $\mathbf{A}^{(l+1)}$ for the next GCN are obtained by equations:
$$\mathbf{X}^{(l+1)}=\mathbf{S}^{(l)\mathrm{T}} \mathbf{Z}^{(l)},$$
$$\mathbf{A}^{(l+1)}=\mathbf{S}^{(l)\mathrm{T}} \mathbf{A}^{(l)}\mathbf{S}^{(l)}.$$

Finally, the representation of the $i$-th graph is defined as $\mathbf{z}_i$, which is the max-pooling of the $\mathbf{X}^{(L)}$.
For clearance, the feed-forward procedure of the hierarchical embedding network used in this paper is summarized in Algorithm \ref{A.ffp}.

\begin{algorithm} [t]\label{A.ffp}\small
	\caption{The feed-forward procedure of the hierarchical GCNs with $L$ DiffPool modules.}
	\LinesNumbered
	\KwIn{\\
		$G_i=(\mathbf{A}_i, \mathbf{X}_i)$ $\leftarrow$ A given graph.\\
		$L\leftarrow$ The number of GCNs / The number of pooling.}
	\KwOut{\\
		$\mathbf{z}_i \in \mathbb{R}^d$: The representation of the graph.}
	
	$\mathbf{X}^{(0)}\leftarrow \mathbf{X}_i$;\\
	$\mathbf{A}^{(0)}\leftarrow \mathbf{A}_i$;\\
	\For{$l=0$ to $L-1$}
	{
		$\mathbf{Z}^{(l)}\leftarrow\mathcal{F}^{(l)}_{embed}(\mathbf{A}^{(l)}, \mathbf{X}^{(l)})$;\\
		$\mathbf{S}^{(l)}\leftarrow softmax_r\left(\mathcal{F}^{(l)}_{pool}(\mathbf{A}^{(l)}, \mathbf{X}^{(l)})\right)$;\\
		$\mathbf{X}^{(l+1)}\leftarrow\mathbf{S}^{(l)\mathrm{T}} \mathbf{Z}^{(l)} \in \mathbb{R}^{n_{l+1}\times d}$;\\
		$\mathbf{A}^{(l+1)}\leftarrow\mathbf{S}^{(l)\mathrm{T}} \mathbf{A}^{(l)}\mathbf{S}^{(l)} \in \mathbb{R}^{n_{l+1}\times n_{l+1}}$;\\
	}
	$\mathbf{z}_i=Maxpool_r(\mathbf{X}^{(L)})$;\\
	
	return $\mathbf{z}_i$;
	
\end{algorithm}

\subsubsection{Binary encoding with Hash functions}
In our method, the network is used to learn representations that are effective for data retrieval. To ensure the framework is applicable to practical large-scale pathological database, we modified the output of the hierarchical GCNs. Letting $\mathbf{Z}=(\mathbf{z}_1^{\mathrm{T}},\mathbf{z}_2^{\mathrm{T}},...,\mathbf{z}_N^{\mathrm{T}})^{\mathrm{T}}\in \mathbb{R}^{N\times d_h}$ represents the graph representations with $N$ denoting the number of training graphs, the hashing function is defined by equation
$$\mathbf{Y}=\tanh(\mathbf{Z}\mathbf{W}_h+\mathbf{b}_h),$$
where $\mathbf{W}_h \in \mathbb{R}^{d_g\times d_h}$ and $\mathbf{b}\in \mathbb{R}^{d_h}$ are the weights and bias for a linear projection and $d_h$ is the dimension of binary codes. $\mathbf{Y}\in(-1,1)^{N\times d_h}$ is the network outputs that can be simply converted into binary codes by equation $$\mathbf{B}=sign(\mathbf{Y})\in\{-1,1\}^{N\times d_h}$$
The loss function to minimize for training the GCN-Hash is defined as 
\begin{equation}\label{E.Hash}
J=\frac{1}{N}\|\frac{1}{d_h} \mathbf{Y}\mathbf{Y}^{\mathrm{T}}-\mathbf{C}\|_F^2 + \lambda\|\mathbf{W}^{\mathrm{T}}\mathbf{W}-\mathbf{E}\|_F^2
\end{equation}
where $\mathbf{C}\in\{-1,1\}^{N\times N}$ is the pair-wise label matrix in which $c_{ij}=1$ represents the $i$-th graph and the $j$-th graph are relevant and $c_{ij}=-1$ otherwise. $\lambda$ is the weight coefficient of the orthogonal regularization. Finally, the proposed GCN-Hash structure is trained end-to-end from the input graph with CNN-features to the output $\mathbf{Y}$. For simplify, the GCN-Hash network is represented as $\mathcal{F}_{GCN-Hash}$.

\subsection{Retrieval using binary codes}
For each WSI in the retrieval database, a set of binary codes ($\mathbf{B}$) that represent the graphs in the WSI can be obtained using the trained feature extraction model $\mathcal{F}_{CNN}$ and GCN-Hash model $\mathcal{F}_{GCN-Hash}$. When retrieving, the region the pathologist queries is divided into patches and converted into binary code using the same model. Then, the similarities between the query code and those in the database are measured using Hamming distance \cite{Zhang2015Towards,zhang2016large-scale,shi2018pairwise,zheng2018histopathological}. After ranking the similarities, the top-ranked regions are retrieved and finally returned to the pathologist. 

\section{Experiments} \label{S.experiment}
\subsection{Experimental setting}
The experiments were conducted on two public datasets of hitopathological whole slide images. The profiles of the datasets are provided as follows.
\begin{itemize}
	\item[$\bullet$] \textit{Camelyon16} \footnote{https://camelyon16.grand-challenge.org} \cite{bejnordi2017diagnositic} contains 400 H\&E-stained lymph node WSIs of breast involving metastases, in which 270 WSIs are defined as training samples and the remainder are used for testing. Regions with cancer in these WSIs are annotated by pathologists. In the evaluation, The training WSIs were used to train the model and establish the retrieval database and the testing WSIs were used to generate query regions.
	\item[$\bullet$] \textit{ACDC-LungHP} \footnote{The dataset is accessible at https://acdc-lunghp.grand-challenge.org. Since the annotations of testing part of the data set are not yet published, only the 150 training WSIs of the data were used in this paper.}~\cite{li2018computer} contains 150 WSIs within lung cancer regions annotated by pathologists. In the evaluation, 30 WSIs were randomly selected as the testing dataset (to generate query regions) and the remainders were used to train the retrieval models and establish the retrieval database.
\end{itemize}

All the WSIs are divided into square patches following the sliding window paradigm. The step of the window was set half of the length of the patch side. DenseNet~\cite{huang2017densely} was employed as the CNN structure to extract patch features. The global average pooling (GAP) layer of the DenseNet structure was used as the feature extractor. The patch size was set $224\times224$ to fit the input of DenseNet. Graphs were constructed for each WSI using the algorithm provided in Algorithm \ref{A.tgc}. For convenience, the graphs for establishing the retrieval database are represented as a set $\mathcal{D}$ and the query graphs are correspondingly represented as $\mathcal{Q}$.

We first conducted experiments to determine hyper-parameters of models involved in our method on the training set. Then, the retrieval performance was evaluated on the testing set and compared with the state-of-the-art methods. 

In the evaluation, the graphs that contain more than 10\% cancerous pixels referring to the pathologists' annotations were defined as \textit{Cancerous Graph}, the graphs containing none cancerous pixels were regarded as \textit{Cancer-free Graph} and the remainders were not counted in the evaluation. Letting $r_{ik}\in\{0,1\}$ represent whether the $k$-th returned result is relevant to the $i$-th query instance. Specifically, $r_{ik}=1$ indicates that the $k$-th result shares the same label with the $i$-th query graph and $r_{ik}=0$, otherwise. The average precision of retrieval $AP(k)$ for top-$k$-returned regions and the mean average precision $mAP$ are used as the metrics, which are defined by equations
$$AP(k)=\frac{1}{|\mathcal{Q}|}\sum_{i=1}^{|\mathcal{Q}|} p_i(k),$$
$$mAP=\frac{1}{|\mathcal{Q}|}\sum_{i=1}^{|\mathcal{Q}|} \frac{\sum_{k=1}^{|\mathcal{D}|} p_i(k)\cdot r_{ik}}{\sum_{k=1}^{|\mathcal{D}|}r_{ik}},$$
where $|\cdot|$ denotes the number of set elements and
$$p_i(k)=\frac{\sum_{j=1}^k{r_{ij}}}{k}$$
is the retrieval precision of the $i$-th query instance for the top-$k$ returned results. The higher the metrics, the better the retrieval performance.

All the experiments were conducted in python with pytorch and run on a computer with double Intel Xeon E5-2670 CPUs, 128 Gb RAM and 4 GPUs of Nvidia GTX 1080Ti.

The models in the proposed method were implemented with pytorch and run on GPUs. The Adam optimizer was employed to train the networks. The dropout with a probability of 0.5 was performed to the neurons of GCNs to relieve over-fitting. 

\subsection{Hyper-parameter determination}
The hyper-parameters of the CNN model and GCN model were determined within the training set. The approaches to determine the hyper-parameters are the same on the two experiment datasets and in this section the detail for the ACDC-LungHP dataset is presented.

\subsubsection{The structure of feature extraction CNN}
The CNN was trained with the patches of the training WSIs via binary classification task. Specifically, patches in size of $224\times224$ pixels were sampled from the slides to train the network, where patches containing above 75\% percentage of cancerous pixels were labeled as positive samples, the patches involving none cancerous pixels were regarded as negative samples and the other patches were not used. In the training, 30\% patches in the training set were spared for validation. 

The depth of DenseNet was tuned within the scope suggested in \cite{huang2017densely}. The best depth was determined according to the classification error of the validation data. Table \ref{T.Densenet} presents the classification results after 20 epochs of training. Lower classification error indicates more discriminative images features. Therefore, the depth of the CNN was determined as 121 according the validation error, for which the dimension of the patch features ($\mathbf{x}_i$) is 1024.

\begin{table} \footnotesize 
	\caption{Classification error for different structure of DenseNet.} \label{T.Densenet}
	\centering
	\begin{tabular}{l | c c c c }
		\hline
		\textbf{Structure} & Train loss & Train error & Valid loss & Valid error \\
		\hline
		\hline
		DenseNet-121 & 0.386 & 0.176 & \textbf{0.363} & \textbf{0.158} \\
		DenseNet-169 & 0.392 & 0.179 & 0.397 & 0.184 \\
		DenseNet-161 & 0.390 & 0.179 & 0.379 & 0.168 \\
		DenseNet-201 & \textbf{0.379} & \textbf{0.172} & 0.385 & 0.165 \\
		\hline
	\end{tabular}
\end{table}

\subsubsection{The structure of GCN-Hash}
\begin{figure*} [t]
	\centering
	\begin{minipage}{0.33\linewidth}
		\subfigure[Number of pooling $L$]{\includegraphics[width=\textwidth]{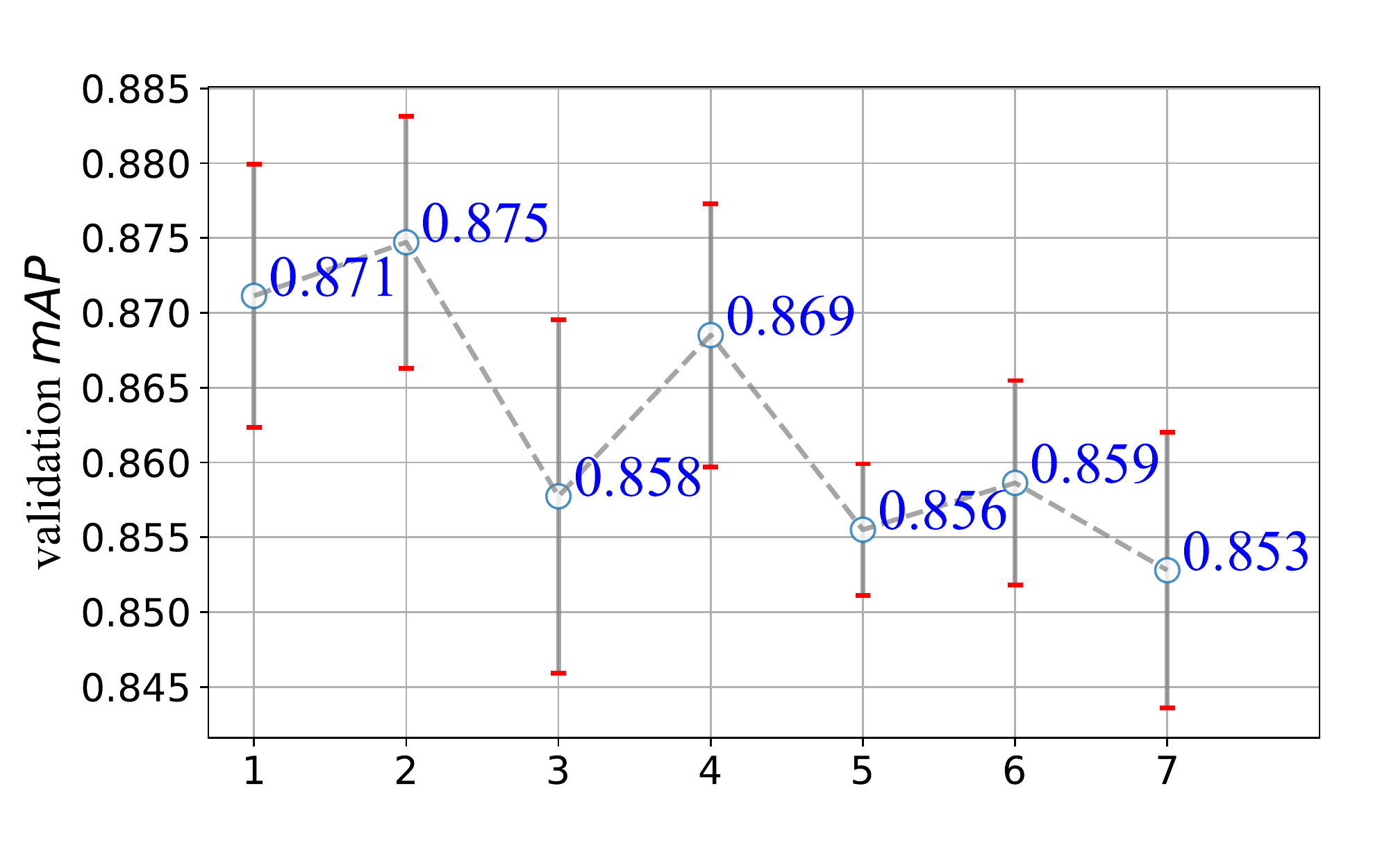}}
	\end{minipage}
	\begin{minipage}{0.33\linewidth}
		\subfigure[Number of graph embedding $K$]{\includegraphics[width=\textwidth]{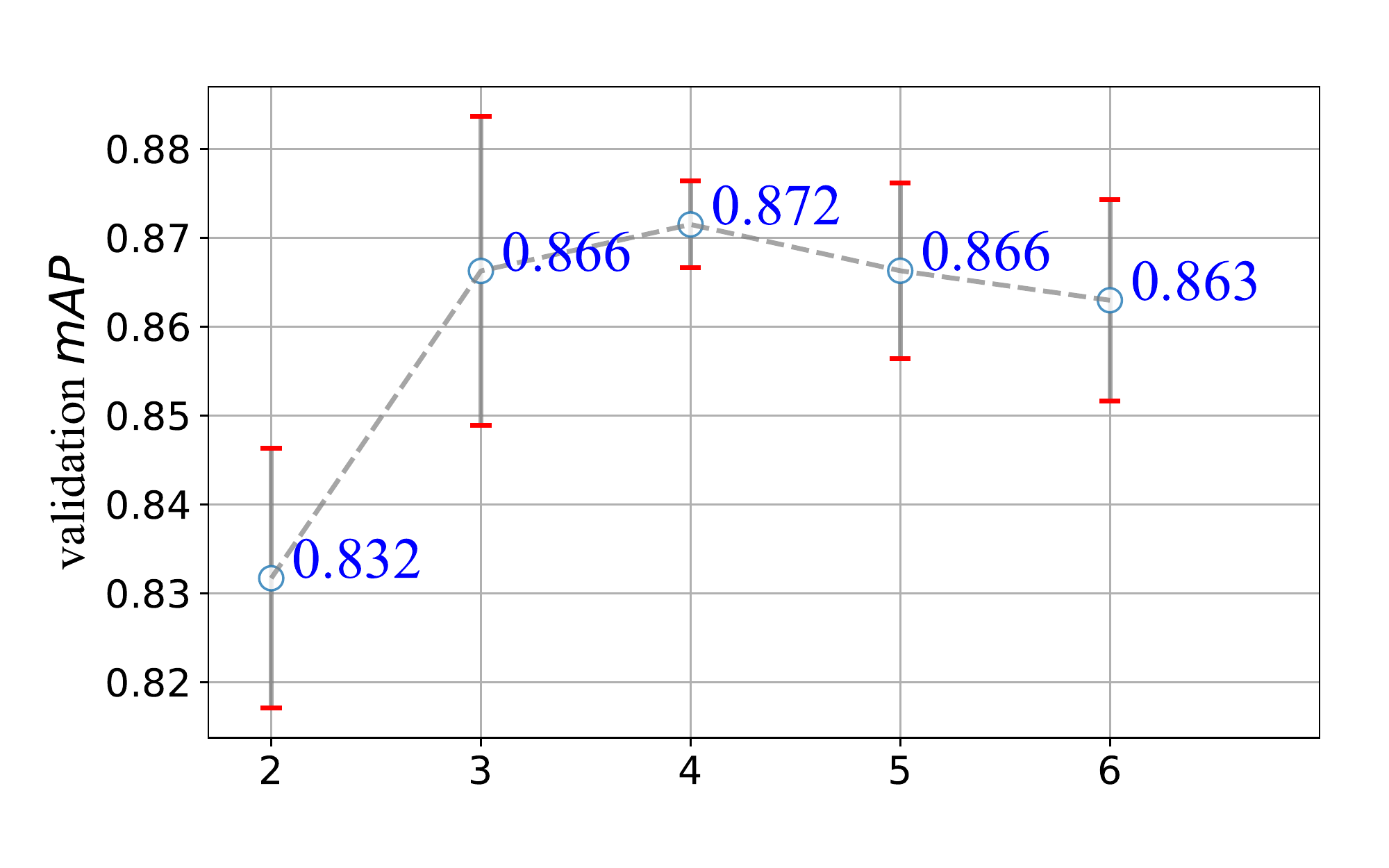}}
	\end{minipage}
	\begin{minipage}{0.33\linewidth}
		\subfigure[Dimension of graph embedding $d$]{\includegraphics[width=\textwidth]{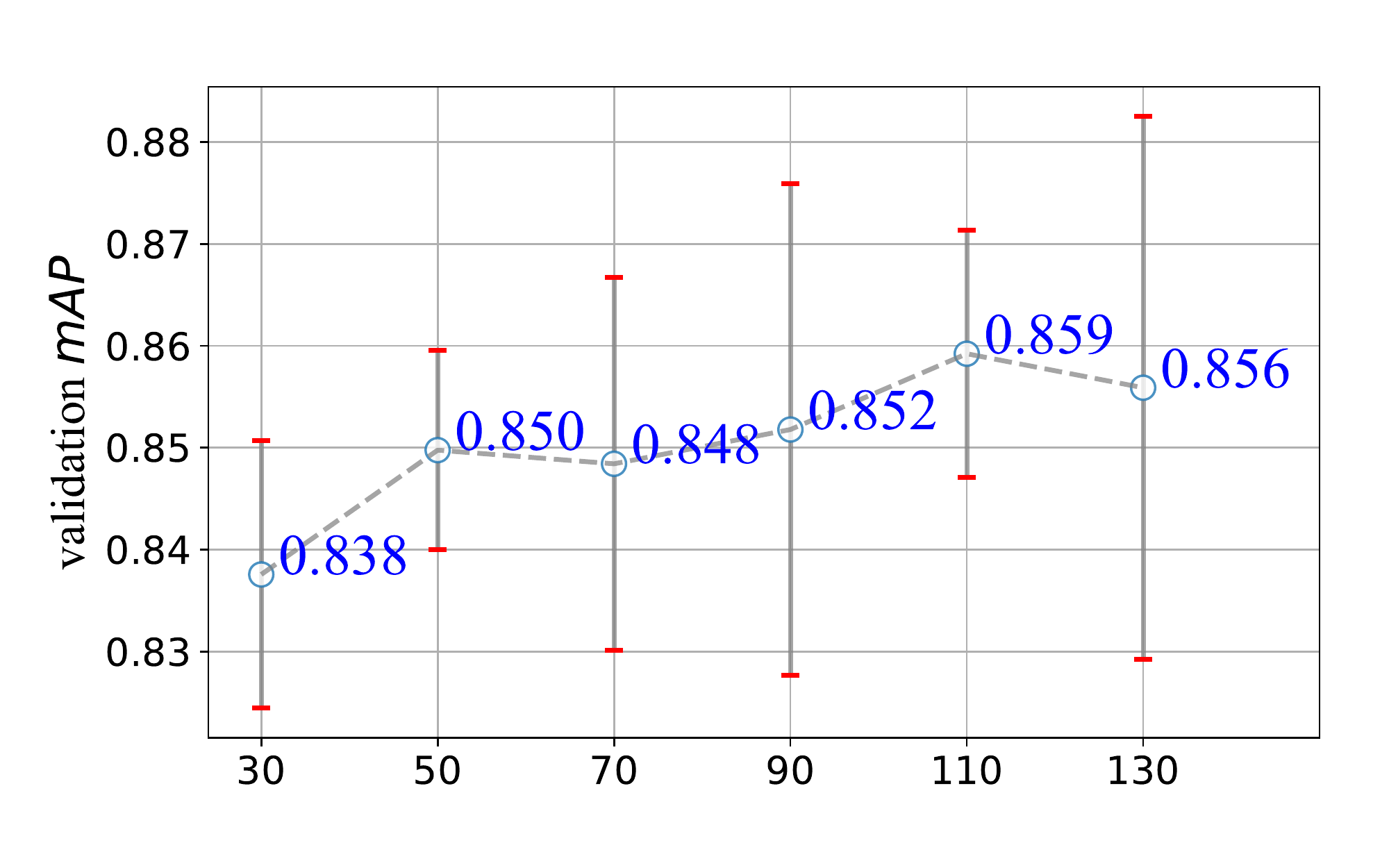}}
	\end{minipage}
	\begin{minipage}{0.33\linewidth}
		\subfigure[Ratio of node reduction $\alpha$ (\%)]{\includegraphics[width=\textwidth]{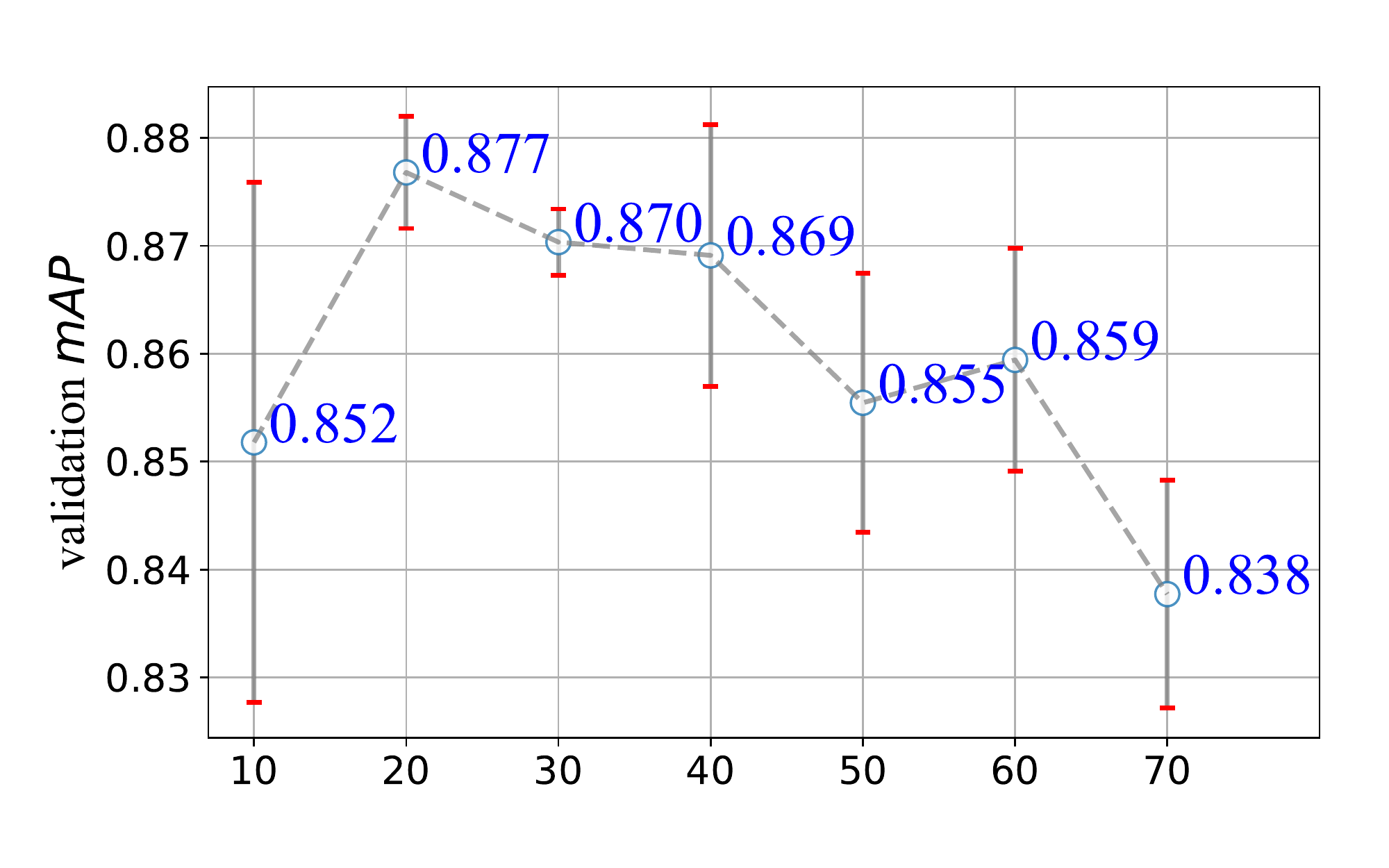}}
	\end{minipage}
	\begin{minipage}{0.33\linewidth}
		\subfigure[Weight in the loss function $\lambda$]{\includegraphics[width=\textwidth]{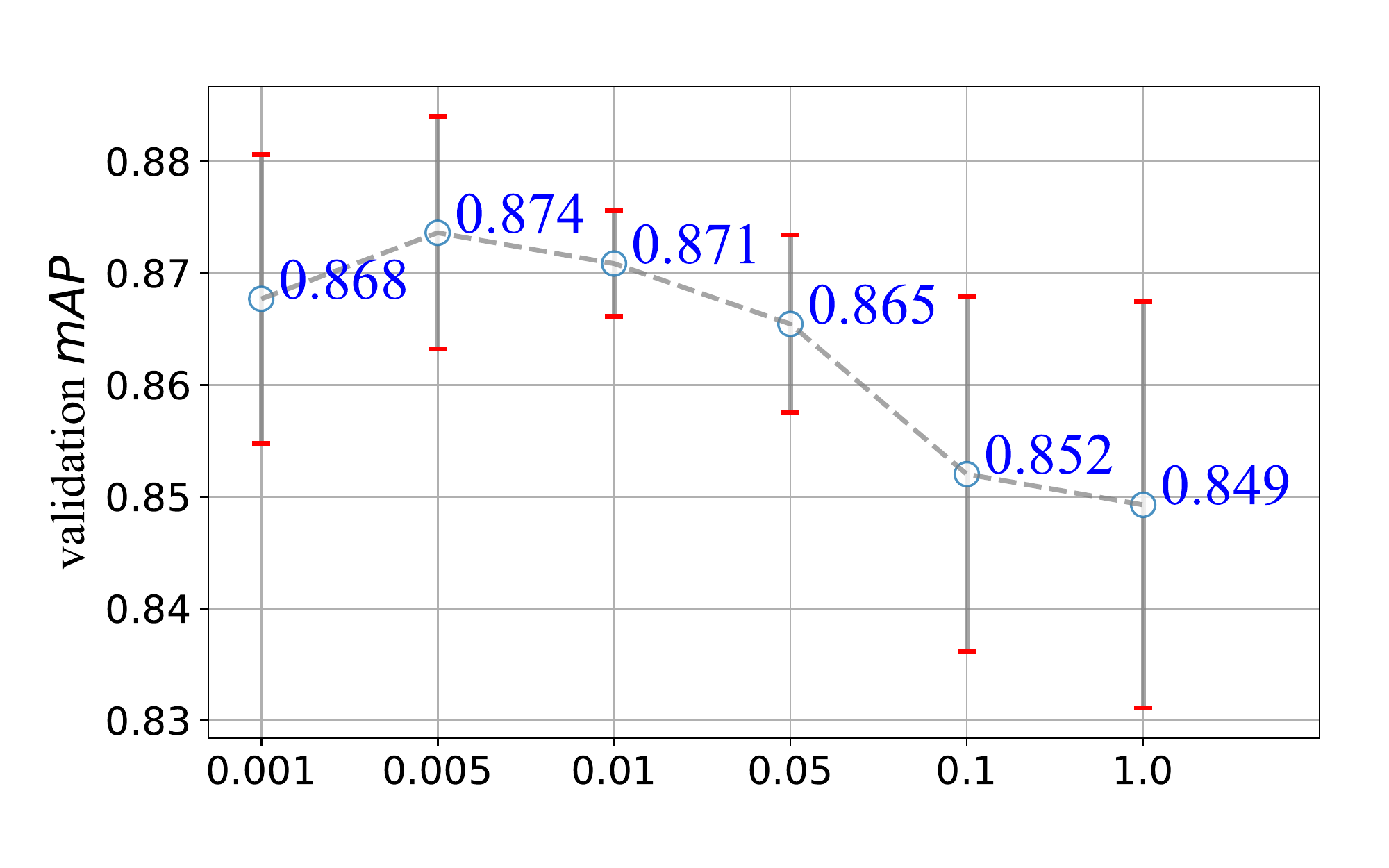}}
	\end{minipage}
	\begin{minipage}{0.33\linewidth}
		\subfigure[Bit number of the hash code $d_h$]{\includegraphics[width=\textwidth]{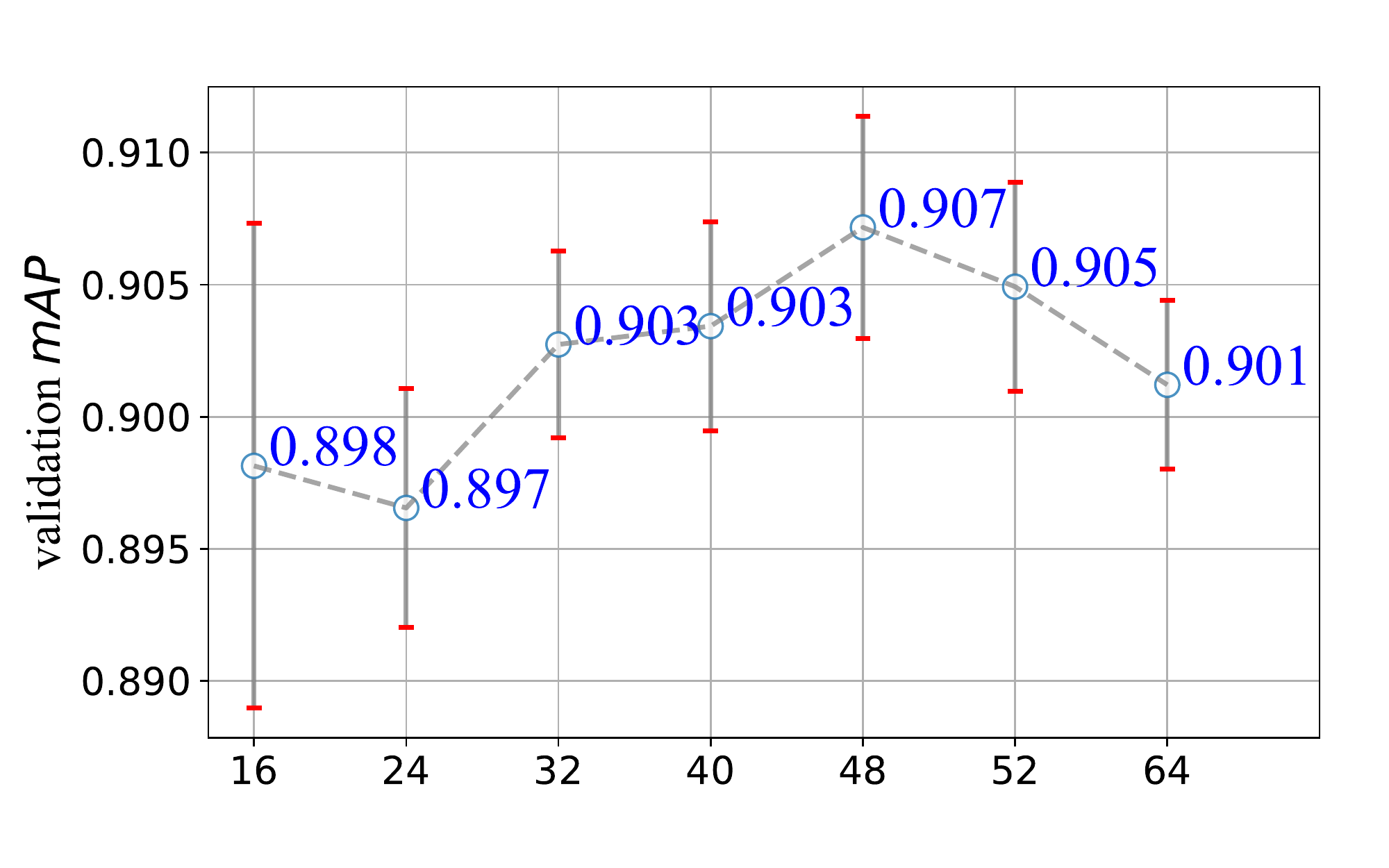}}
	\end{minipage}
	\caption{The $mAP$ curves as functions of the hyper-parameters of the GCN-Hash model, where the average $mAP$ of the 5-fold cross-validation is presented by each data point and the standard variance of the 5 trials is drawn with red bar.} \label{F.hp}
\end{figure*}

There are 6 hyper-parameters in the proposed GCN-Hash model. These parameters were tuned in large ranges and determined based the best mAP obtained through 5-fold cross-validation in the training set, where for each fold one part were used as query graphs and the other four parts were used to train the model and establish the database. Note that the other hyper-parameters were set fixed when one hyper-parameter was tuned. The experimental results are presented in Fig. \ref{F.hp}. 

1) The depth of the network is determined by the number of GCNs (or DiffPool modules) $L$ and the step of embedding $K$ in each GCN. A larger $L$ helps extract the local information in hierarchical way but would increase the risk of over-fitting. Therefore, $L$ was tuned from 1 to 7 and the retrieval metrics (as shown in Fig. \ref{F.hp} (a)) indicate $L=2$ is optimum for the dataset. In GCN, the contextual information considered for node encoding enlarges as the step of embedding $K$ increases. Particularly for histopathology diagnosis, the allocation of tissue objects is important indication for cancer diagnosis. Therefore, we ranged $K$ from 2 to 6 to find an appropriate receptive field of node on the graph. According to the results in Fig. \ref{F.hp} (b), $K$ was set to 4.

2) The dimension of graph embedding $d$ (defined in Eq. \ref{E.graph_embed}) determines the \textit{width} of the network after $L$ and $K$ are fixed. $d$ is positive correction with the number of trainable weights involved in the network. Besides, the ratio of nodes reduction after each DiffPool module also related to the number of trainable weights. The ratio is defined as $\alpha=n_{l+1}/n_l$ which controls the speed of nodes clustering. The mAP curves for different settings of the two parameters are provided in Fig. \ref{F.hp} (c-d). According to the results, $d$ was set 110 and $\alpha$ was set 0.2 in the following experiment.

3) The loss function (Eq. \ref{E.Hash}) involves two main hyper-parameters: The weight coefficient of the orthogonal regularization $\lambda$ and the bit number of the binary code $d_h$. The $\lambda$ was tuned in the log space and was set to 0.005 according to the result of Fig. \ref{F.hp} (e). The code space of Hash is in direct proportion to $2^{d_h}$. In practical application, $d_h$ should be set according to the amount of information in the retrieval database. $d_h$ was validated from 16 to 64, for which the results are shown in Fig. \ref{F.hp} (f). The curve indicates that 48 bit is adequate to cover the information of the database. Therefore, it was set $d_h=48$ in the following experiment.

\subsection{Comparison with the state-of-the-art}
The proposed method is compared with 4 state-of-the-art-methods \cite{shi2018pairwise,ma2018generating,jim2017deep,zheng2018size-scalable} proposed for histological image retrieval, where the first method is designed for image patch retrieval and the others are proposed for whole slide image applications. The implementation details of the compared methods are briefly introduced as follows. For convenience, the irregular regions for retrieval are all referred by graphs. 

\begin{itemize}
	\item[$\bullet$] \cite{shi2018pairwise} The patches are fed into an end-to-end hashing networks based on CNN backbones to generate binary codes. Since the method was developed for database containing patches with fixed size, we modified it to adapt the irregular region retrieval. Specifically, the minimum distance between patches across two graphs was used as the distance between the two graphs.
	\item[$\bullet$] \cite{jim2017deep} The retrieval is achieved based on both the WSIs and the text information of the cases. In the experiment, only the part for WSIs retrieval was implemented for the meta-information of the datasets is not available. Specifically, the distances between all pairs of patch features across two graphs were calculated and the mean value of the distances was used as the similarity measurement.
	\item[$\bullet$] \cite{ma2018generating} The features involved in an tissue graph are quantified through max-pooling operation. Then, the obtained representations are converted into binary codes based on latent Dirichlet allocation (LDA) \cite{blei2003latent} followed by supervised hashing. Finally, the similarity between two graphs is computed based on binary codes.
	\item[$\bullet$] \cite{zheng2018size-scalable} The patches in the graphs are encoded into binary codes. When retrieving, a set of proposal graphs are first retrieved through table lookup operation based on patch codes. Then, the distances between query graph and the proposal graphs are calculated under specific similarity measurement and then the most similar graphs are returned.
\end{itemize}

For fair comparison, the feature extractors (or backbones) of the compared methods were the same DenseNet-121 structure.

The GCN and the Hash function are two essential components in our model. Therefore, we conducted ablation experiments to verify the performance of the two components, for which the implementations of the ablation models are described as follows.

\begin{itemize}
	\item[$\bullet$] The proposed w/o link: The links among nodes in the graph are removed by setting the adjacency matrix $\mathbf{A}_i=\mathbf{0}$. Then, the embedding functions of GCN defined in Eq. \ref{E.gcn} is degraded as node-wise neural networks.
	\item[$\bullet$] The proposed w/o Hash: The loss function (Eq. \ref{E.Hash}) is replaced by cross-entropy function with softmax outputs, which is used for graph classification in \cite{ying2018hierarchical}. Then, the last embedding of the network is regarded as the graph code and used to calculate the similarities. The cosine distance is employed as the similarity measurement.
\end{itemize}
\begin{table*} [!ht]\footnotesize
	\caption{Retrieval performance for the state-of-the-art methods, where the results for different size allocations of graphs (determined by $\bar{n}$) are compared.}\label{T.results1}
	\centering
	\begin{tabular}{l|ccccccc}
		\hline \multirow{2}{1cm}{\textbf{Methods}} & {$\bar{n}=30$}  & {$\bar{n}=40$}  & {$\bar{n}=50$}  & {$\bar{n}=60$}  & {$\bar{n}=70$}  & {$\bar{n}=80$}  & {$\bar{n}=90$} \\
		\cline{2-8}
		& $AP(50)$/mAP  & $AP(50)$/mAP  & $AP(50)$/mAP  & $AP(50)$/mAP  & $AP(50)$/mAP  & $AP(50)$/mAP  & $AP(50)$/mAP \\
		\hline
		\hline
		Shi \etal \cite{shi2018pairwise} & 0.748/ 0.671& 0.732/ 0.664& 0.720/ 0.656& 0.714/ 0.651& 0.699/ 0.644& 0.699/ 0.641& 0.699/ 0.638\\
		Zheng \etal \cite{zheng2018size-scalable} & 0.797/ 0.702& 0.788/ 0.704& 0.789/ 0.703& 0.794/ 0.703& 0.780/ 0.701& 0.790/ 0.703& 0.793/ 0.704\\
		Ma \etal \cite{ma2018generating}& 0.783/ 0.715& 0.783/ 0.719& 0.779/ 0.719& 0.788/ 0.718& 0.783/ 0.717& 0.789/ 0.718& 0.786/ 0.718\\
		Jimenez \etal \cite{jim2017deep}& 0.779/ 0.708& 0.777/ 0.709& 0.772/ 0.709& 0.779/ 0.708& 0.765/ 0.705& 0.786/ 0.707& 0.782/ 0.707\\
		\hline
		The proposed w/o Hash & 0.786/ 0.814& 0.792/ 0.843& 0.822/ 0.859& 0.812/ 0.842 & \textbf{0.799}/ 0.819& 0.804/ 0.804& 0.838/ 0.858\\   
		The proposed w/o link & \textbf{0.803}/ 0.851 & \textbf{0.812}/ 0.852& 0.784/ 0.834& 0.823/ 0.867& 0.774/ 0.833& 0.821/ 0.833& 0.828/0.865\\
		The proposed & 0.801 / \textbf{0.862}& 0.811 / \textbf{0.865} & \textbf{0.840} / \textbf{0.867} & \textbf{0.831} / \textbf{0.872}& 0.797 / \textbf{0.857} & \textbf{0.845} / \textbf{0.884} & \textbf{0.858} / \textbf{0.881}\\
		\hline
	\end{tabular}
\end{table*}
\begin{figure*}[!ht]
	\centering
	\begin{minipage}{\linewidth}
		\subfigure[The boxplots of node number distributions for graphs generated with different $\bar{n}$, where the nubmer of graphs is located under each box and the median number is marked by red line.]{\includegraphics[width=0.66\textwidth]{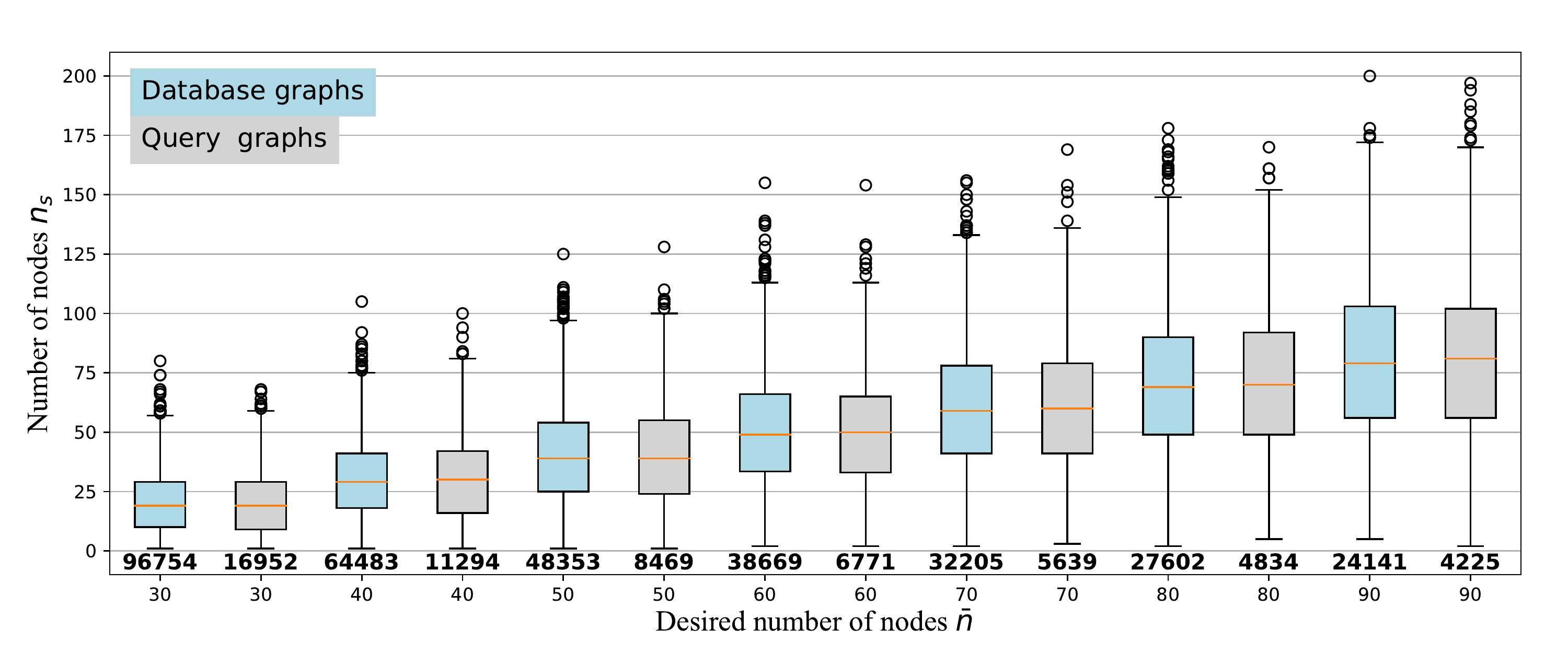}}
		\subfigure[$\bar{n}=30$]{\includegraphics[width=0.33\textwidth]{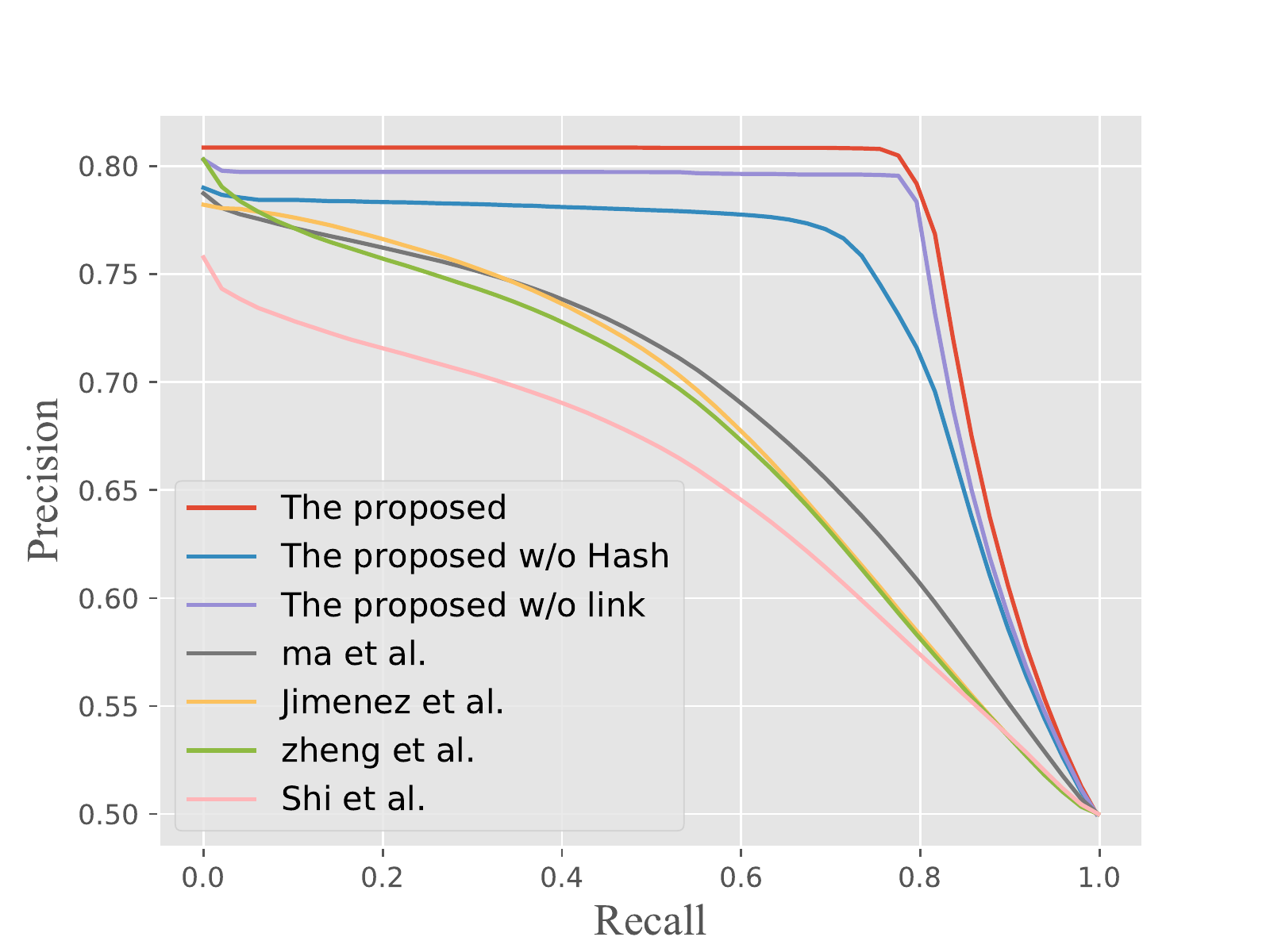}}
		
		\subfigure[$\bar{n}=40$]{\includegraphics[width=0.33\textwidth]{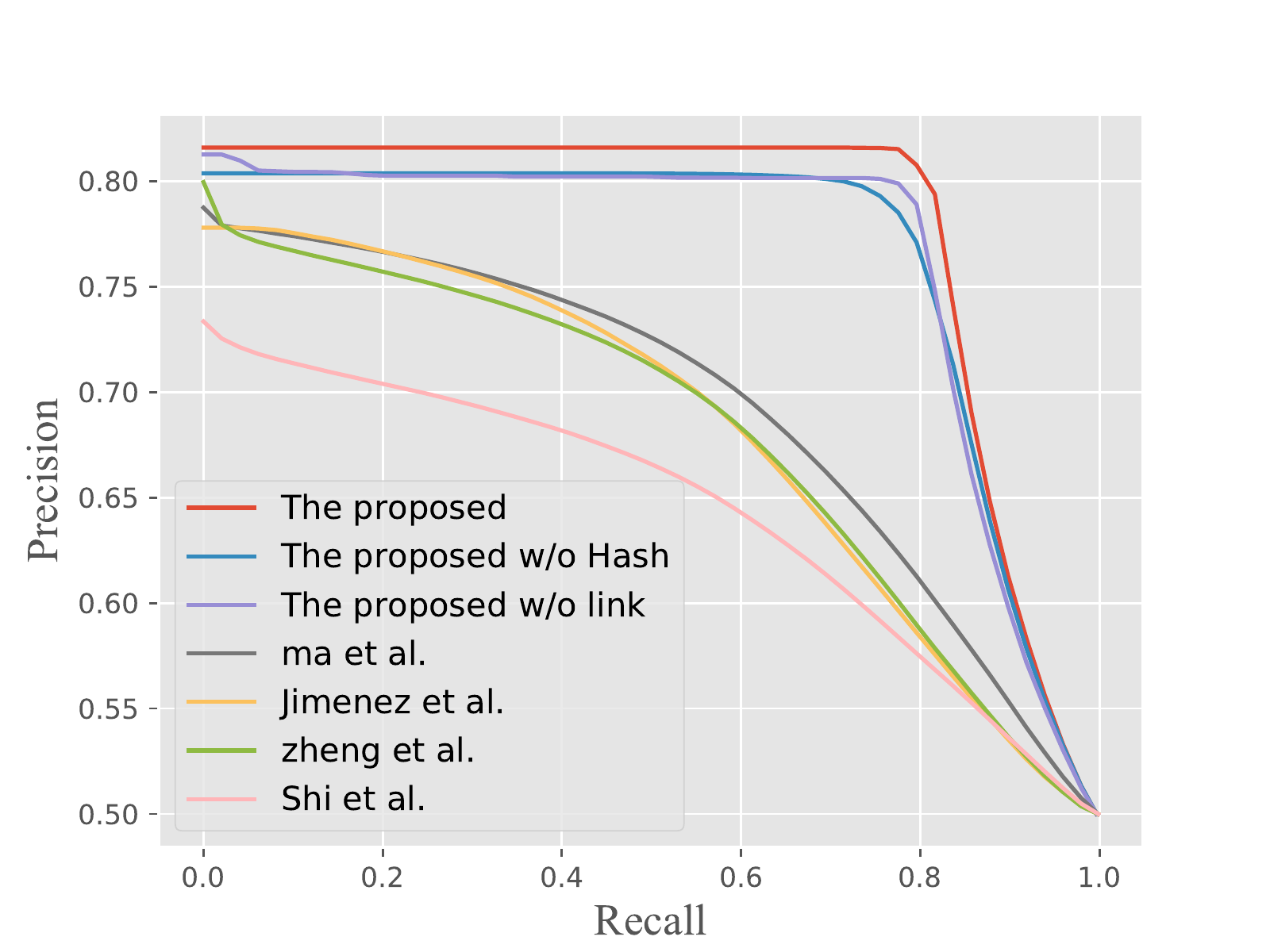}}
		\subfigure[$\bar{n}=50$]{\includegraphics[width=0.33\textwidth]{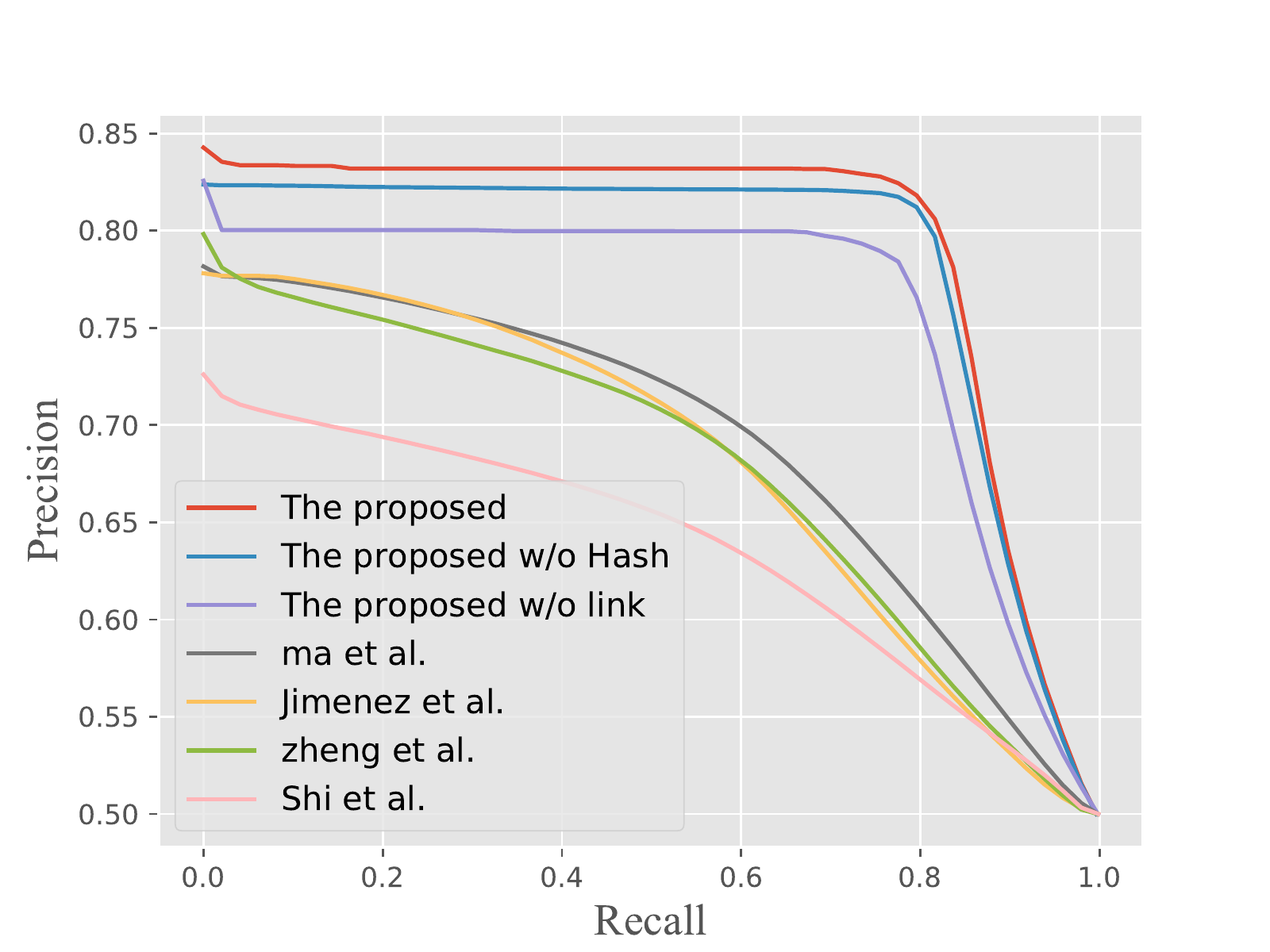}}
		\subfigure[$\bar{n}=60$]{\includegraphics[width=0.33\textwidth]{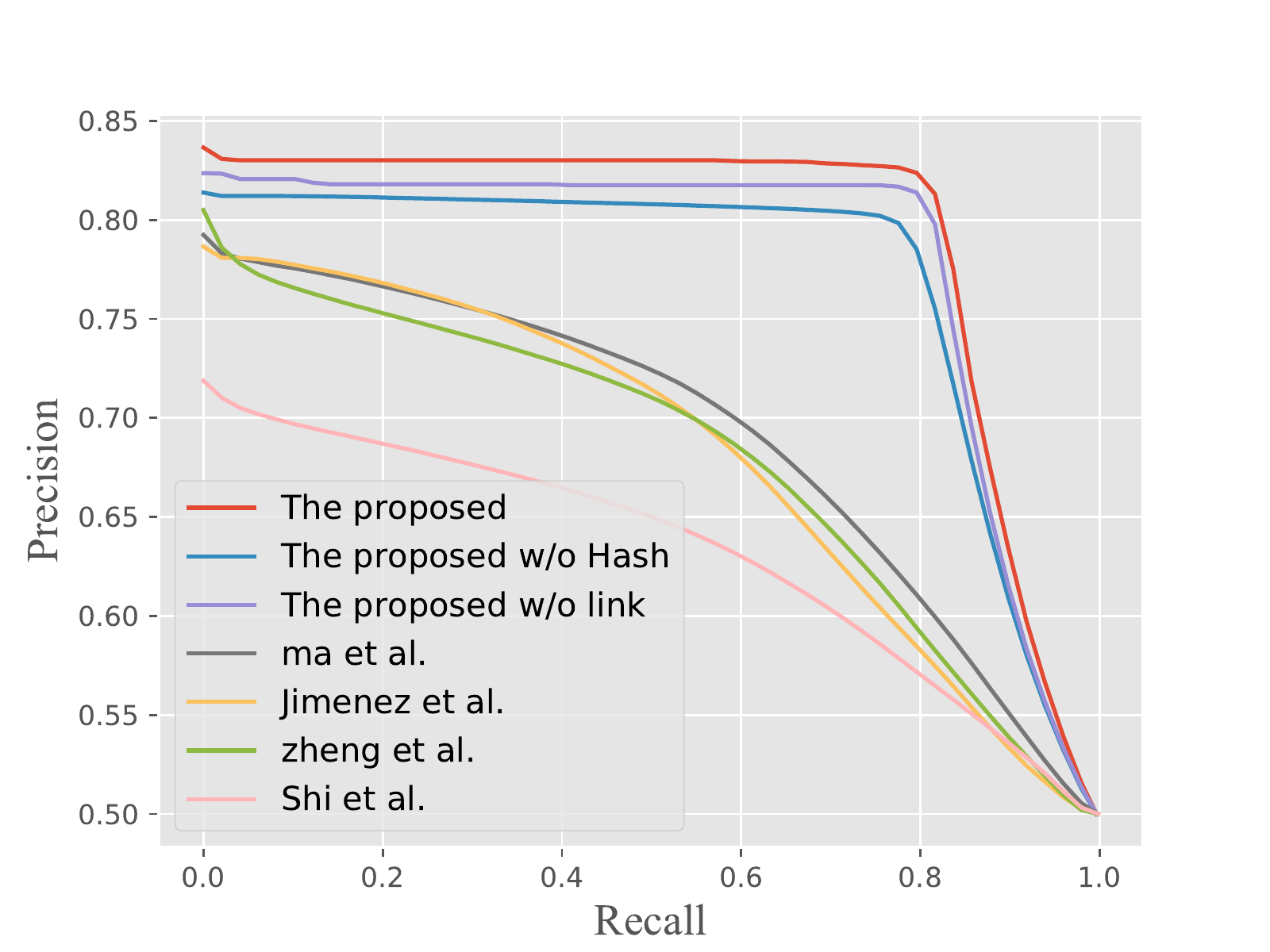}}
		
		\subfigure[$\bar{n}=70$]{\includegraphics[width=0.33\textwidth]{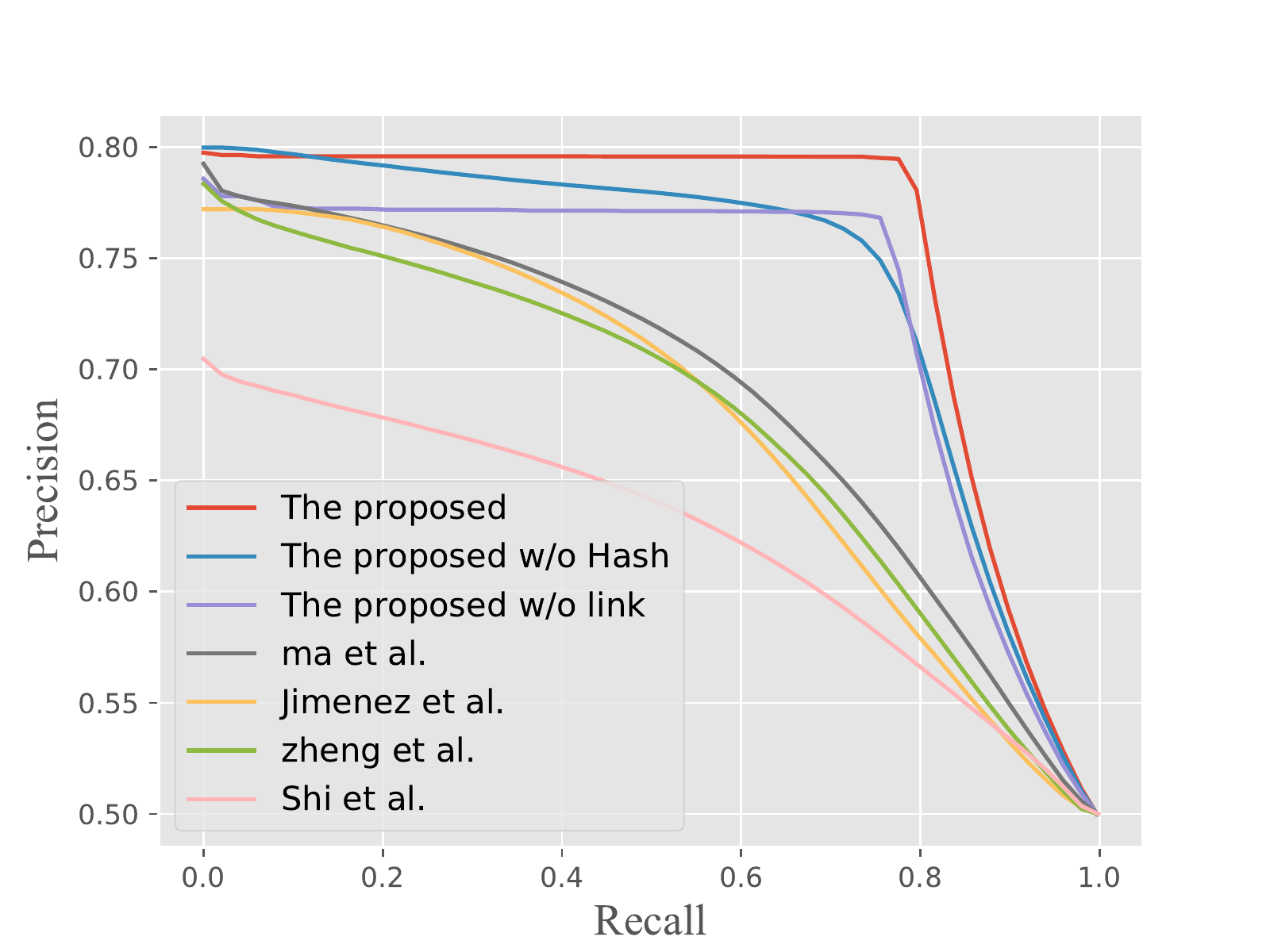}}
		\subfigure[$\bar{n}=80$]{\includegraphics[width=0.33\textwidth]{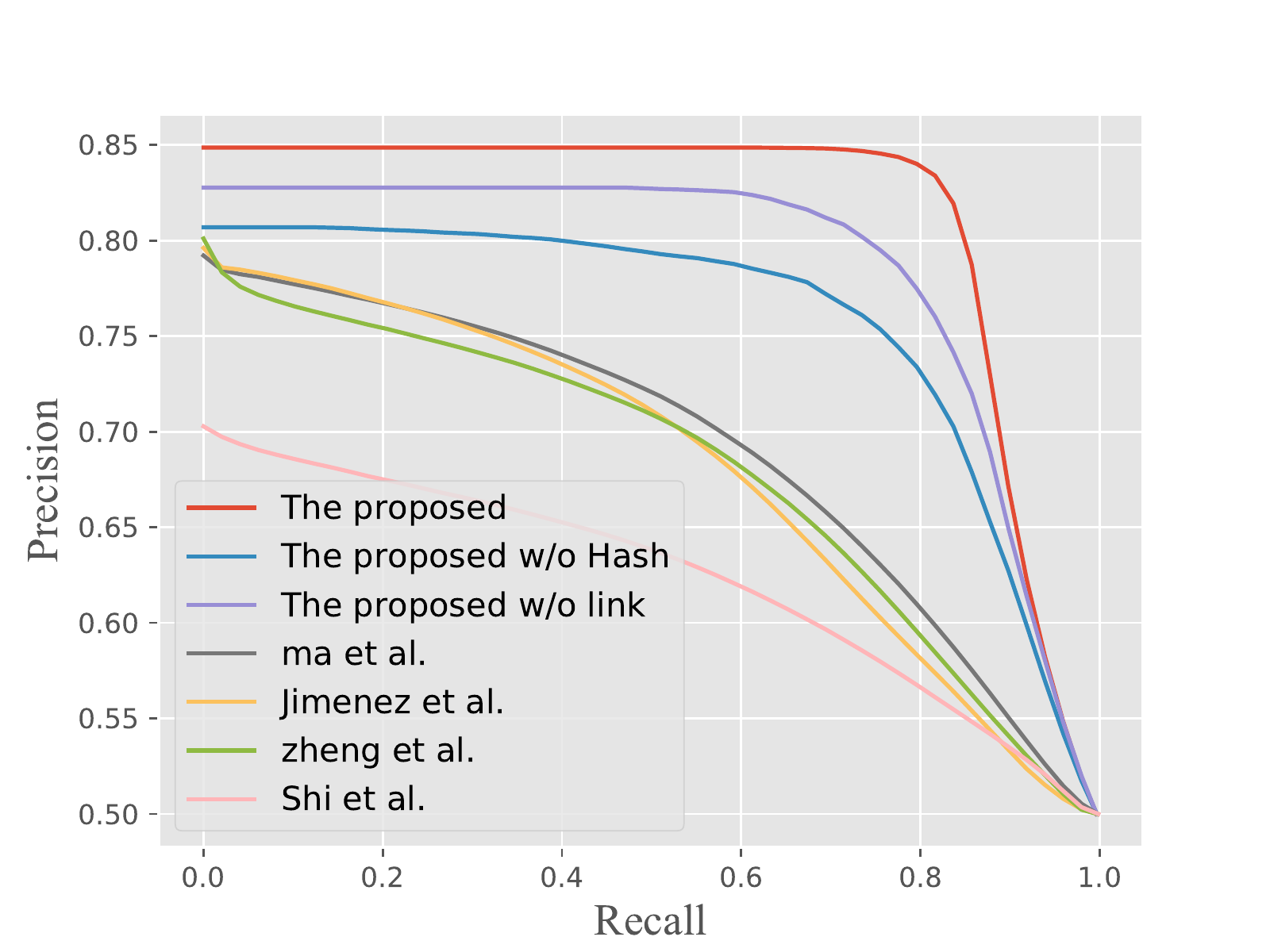}}
		\subfigure[$\bar{n}=90$]{\includegraphics[width=0.33\textwidth]{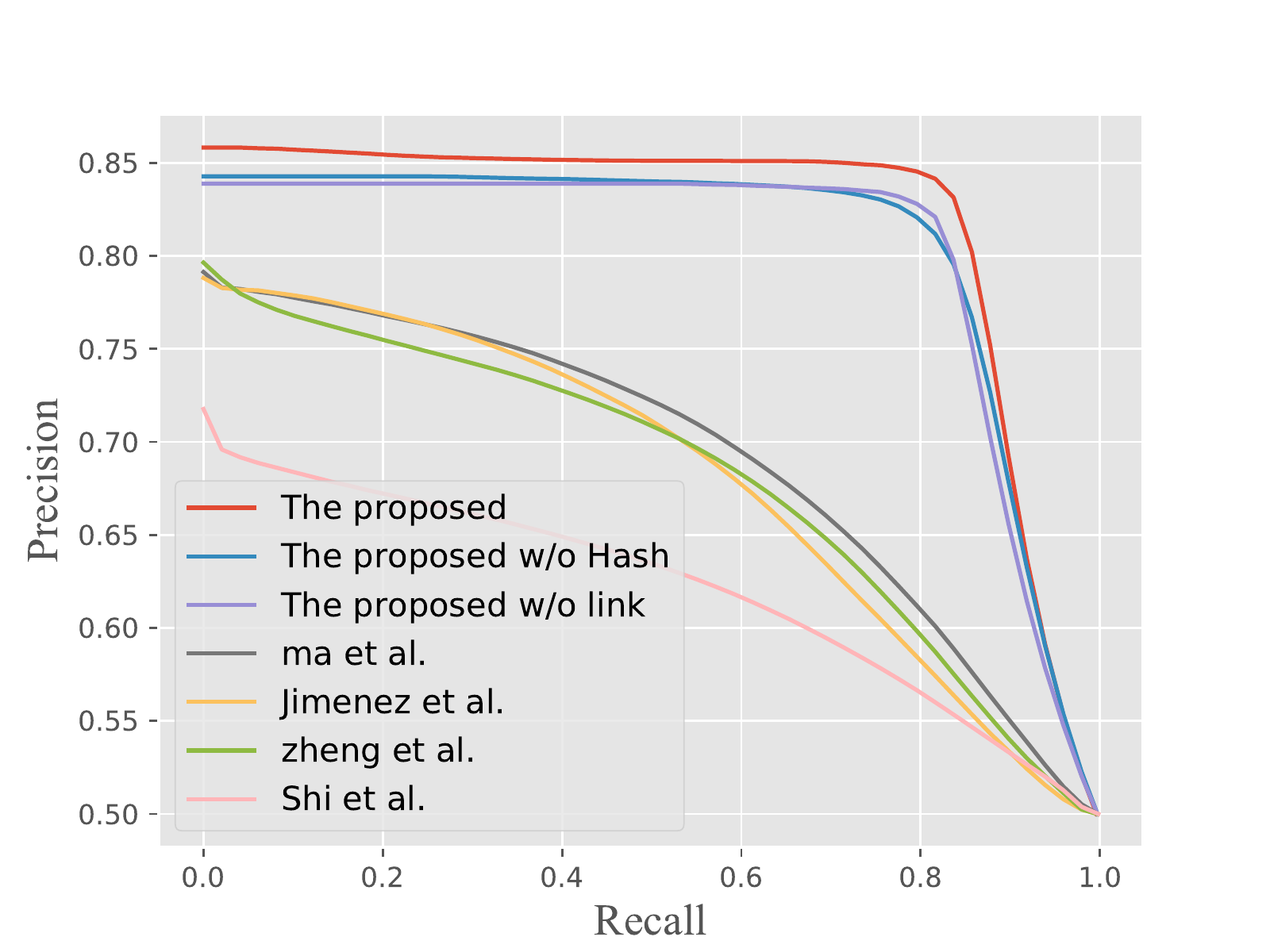}}
	\end{minipage}
	\caption{Comparison of interpolated precision-recall curves of different retrieval methods, where (a) provides the distributions of number of graph nodes obtained with different $\bar{n}$, and (b-h) present the interpolated precision-recall curves for different settings of $\bar{n}$, respectively.} \label{F.pr_curves}
\end{figure*}

\subsubsection{Comparison of retrieval precision}
We conducted experiments with graphs in different scales to comprehensively evaluate the retrieval performance of the compared method. Specifically, the number of clusters (i.e. the target number of graphs $\hat{g}_s$ in Algorithm \ref{A.tgc}) for each WSI is determined by equation $\hat{g}_s=[m_s/\bar{n}]$, where $\bar{n}$ controls the desired number of nodes in each graph. In the comparison, $\bar{n}$ was set from 30 to 90 with a step of 10 and the retrieval database and query graphs for each setting of $\bar{n}$ were obtained based on Algorithm \ref{A.tgc}. The allocation of graph node numbers are presented with boxplots in Fig. \ref{F.pr_curves}. 

The experimental results are summarized in Table~\ref{T.results1}. Correspondingly, the interpolated precision-recall curves are illustrated in Fig. \ref{F.pr_curves} (b-h). Overall, the proposed method has achieved the best retrieval performance in the quantitative evaluation. 


The retrieval strategy in methods~\cite{jim2017deep,ma2018generating,zheng2018size-scalable} were also designed for query regions in various size or shape. 
In \cite{ma2018generating}, the patch features in a region were quantified using max-pooling operation and the similarities between regions were calculated based on the pooled representations. The percentage of patches of different tissue types was ignored by the pooling operation. While, zheng \etal \cite{zheng2018size-scalable} and Jimenez-del-Toro \etal \cite{jim2017deep} proposed measuring the similarity of two regions by an ensemble of feature distances of all the patch pairs across the two regions. The local similarities between the regions are well measured in the two methods, but the global information was discarded. Furthermore, the adjacent relationship of tissue objects cannot be effectively described in these methods. In contrast, the proposed method constructed graphs within tissue regions. The information of tissue allocation has been sufficiently described and well preserved through the hierarchical embeddings of the proposed GCN-Hash model. It contributes to a significant improvement compared to the previous methods. The mAP of retrieval in ACDC-LungHP dataset is above 0.857, which is 14.0\% to 16.3\% higher than the compared results in the task of irregular regions retrieval. 

The results of the ablation experiments are compared in the last three rows of Table. \ref{T.results1}. Overall, the retrieval precision reduced when the Hash model and edges (links) of graph were discarded. Especially for large regions retrieval (e.g. the results obtained under $\bar{n}>50$), the consideration of graph links contributed an improvement of 2.4\%\ to 5.1\% to the $mAP$. The experimental results have demonstrated that 1) the adjacency relationship of tissue object is necessary and significant in the learning of tumor region representations and 2) The Hash function is effective to learn the representations for the irregular regions retrieval from WSI-database.

\begin{table} 
	\caption{Computational complexity for the compared methods.}\label{T.cc}
	\centering
	\begin{tabular}{l|l| l}
		\hline
		\textbf{Methods} & Complexity & Time \\
		\hline
		\hline
		Shi \etal \cite{shi2018pairwise} & $\mathcal{O}(a^2bc)$ & 6.51 s\\
		Zheng \etal \cite{zheng2018size-scalable} & $\mathcal{O}(a^2bc)$ & 91.8 ms\\
		Ma \etal \cite{ma2018generating} & $\mathcal{O}(bc)$ & 0.791 ms\\
		Jimenez-del-Toro \etal \cite{jim2017deep} & $\mathcal{O}(a^2bc)$ & 6.71 s\\
		\hline
		The proposed w/o Hash & \multirow{3}{1cm}{$\mathcal{O}(bc)$} & 3.47 ms\\
		\cline{3-3}
		The proposed w/o link &  & \multirow{2}{1.5cm}{0.802 ms} \\
		The proposed & & \\
		\hline
	\end{tabular}
\end{table}

\subsubsection{Comparison of computational complexity}
The efficiency is equally important for CBHIR system. In the online retrieval stage, the computational complexity mainly derives from the strategy of retrieval, which is relevant to the pixel size of query region $a$, the scale of the database $b$ and the dimension of region representations $c$. The $\mathcal{O}$ notation for the compared methods are given in Table~\ref{T.cc}. Correspondingly, the average time consumptions for the retrieval experiment with $\bar{n}=50$ are compared (the feature extraction time is not involved) and provided in the Table~\ref{T.cc}. 

In our method, the computation for retrieval is irrelevant to the size of query region after the encoding and therefore the complexity is $\mathcal{O}(bc)$ (The computational complexity of Hamming distance is $\mathcal{O}(c)$). The average time for querying the database containing 48,353 graphs is 0.802 ms. Moreover, benefiting from the binary encoding, the similarity measurement is time-saving than those based on float-type high-dimensional features (e.g. The proposed w/o Hash and Jimenez-del-Toro \etal \cite{jim2017deep}). When the order of magnitudes of WSI in database increases and the content in the database is abundant, a Hash table can be pre-established. Then, the retrieval can be easily achieved by a table-lookup operation, for which the complexity of retrieval is potentially reduced to $\mathcal{O}(1)$.

\begin{table*} \footnotesize
	\caption{Retrieval performance on Camelyon16 dataset, where the results for different size allocations of graphs (determined by $\bar{n}$) are compared.}\label{T.results2}
	\centering
	\begin{tabular}{l|ccccccc}
		\hline\multirow{2}{1cm}{\textbf{Methods}} & {$\bar{n}=30$}  & {$\bar{n}=40$}  & {$\bar{n}=50$}  & {$\bar{n}=60$}  & {$\bar{n}=70$}  & {$\bar{n}=80$}  & {$\bar{n}=90$} \\
		\cline{2-8}
		& $AP(50)$/mAP  & $AP(50)$/mAP  & $AP(50)$/mAP  & $AP(50)$/mAP  & $AP(50)$/mAP  & $AP(50)$/mAP  & $AP(50)$/mAP \\
		\hline
		\hline
		Shi \etal \cite{shi2018pairwise} & 0.834/ 0.691 & \textbf{0.842}/ 0.698& 0.847/ 0.708& 0.860/ 0.716& 0.865/ 0.723& 0.869/ 0.724& 0.869/ 0.727\\
		Zheng \etal \cite{zheng2018size-scalable} & 0.828/ 0.688& 0.837/ 0.695& 0.852/ 0.707& 0.867/ 0.715& 0.867/ 0.722& 0.874/ 0.723& 0.874/ 0.727\\
		Jimenez \etal\cite{jim2017deep}  & 0.809/ 0.693& 0.816/ 0.701& 0.827/ 0.717& 0.839/ 0.724& 0.847/ 0.736& 0.847/ 0.737& 0.856/ 0.745\\
		Ma \etal \cite{ma2018generating} & 0.796/ 0.663& 0.806/ 0.667& 0.816/ 0.681& 0.825/ 0.683& 0.830/ 0.691& 0.836/ 0.692& 0.839/ 0.697\\
		The proposed & \textbf{0.862}/ \textbf{0.879}& 0.841/ \textbf{0.864} & \textbf{0.881}/ \textbf{0.897} & \textbf{0.895}/ \textbf{0.911} & \textbf{0.896}/ \textbf{0.909} & \textbf{0.884}/ \textbf{0.909} & \textbf{0.895}/ \textbf{0.906}\\
		\hline
	\end{tabular}
\end{table*}

\subsection{Comparison on Camelyon16 dataset}
The same evaluations were completed on the Camelyon16 dataset. The hyper-parameters of the GCN-Hash model were tuned within the 270 training WSIs and were finally determined as $(L,K,d,\alpha,\lambda,d_h)=(2,4,100,0.2,0.05,48)$. Then the training WSIs were encoded to construct the retrieval database. The 130 testing WSIs were used to generate the query graphs. The metrics of retrieval for different settings of $\bar{n}$ are compared in Table \ref{T.results2}. It shows that the proposed method achieved best retrieval performance. The experiment results are consistent with those obtained on ACDC-LungHP dataset.

\subsection{Visualization}
\begin{figure*}
	\centering
	\includegraphics[width=\linewidth]{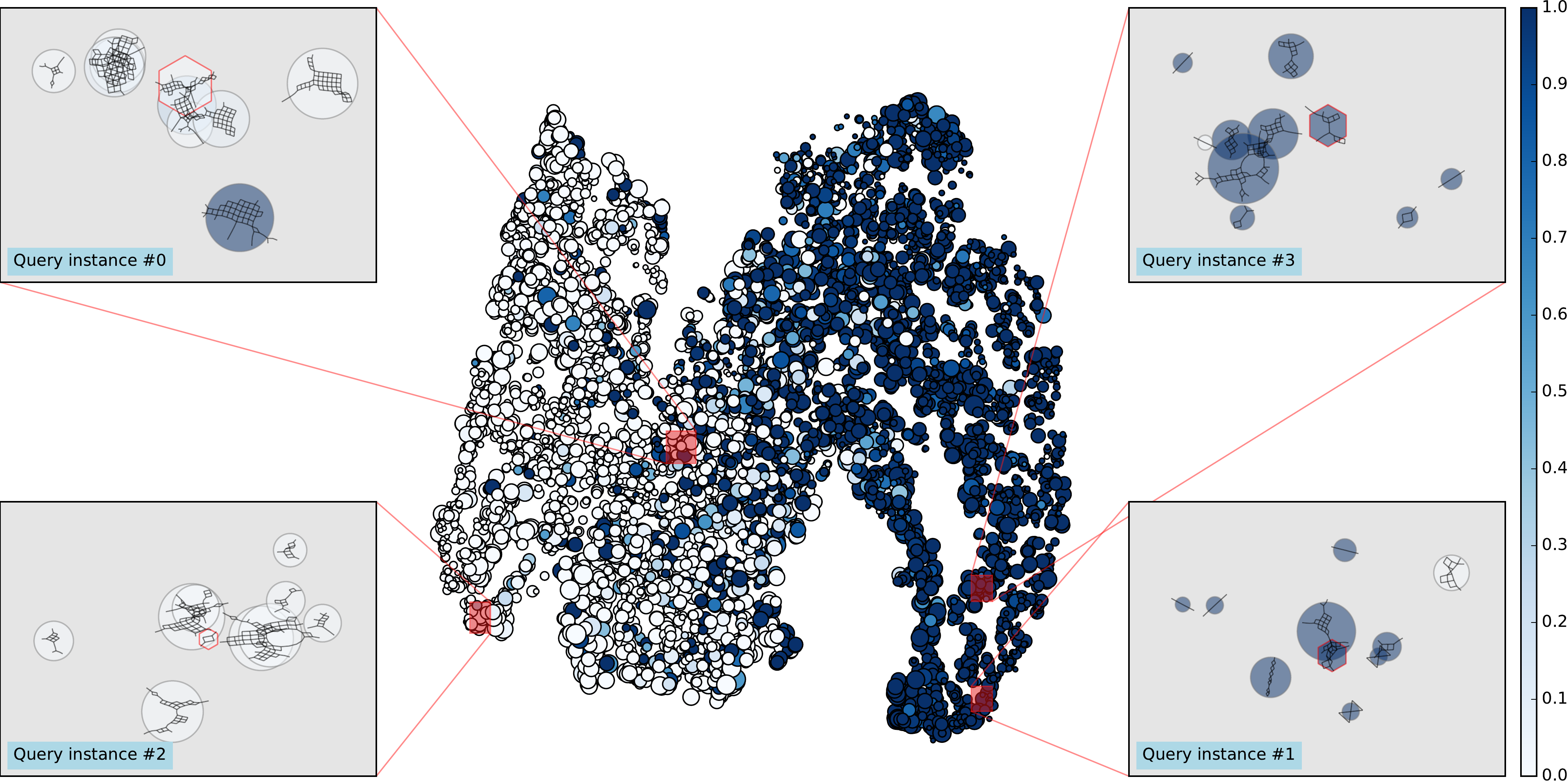}
	\caption{The 2-dimensional visualization of the output of GCN-Hash model ($\mathbf{Y}$) for the retrieval database of ACDC-LungHP, where a dot represents a graph, the body color of the dot indicates the ratio of tumor occupation in the graph referring to the color bar on the right of the figure and the size of the dot indicates the scale of graphs. the zoom-in regions focus on four retrieval instances, where the query graph is symbolized by red hexagon, and the corresponding graph layouts are drawn within the circles. For clear display, only a part of the graphs that selected through randomly sampling are plotted.} \label{F.tsne}
	
	\includegraphics[width=\textwidth]{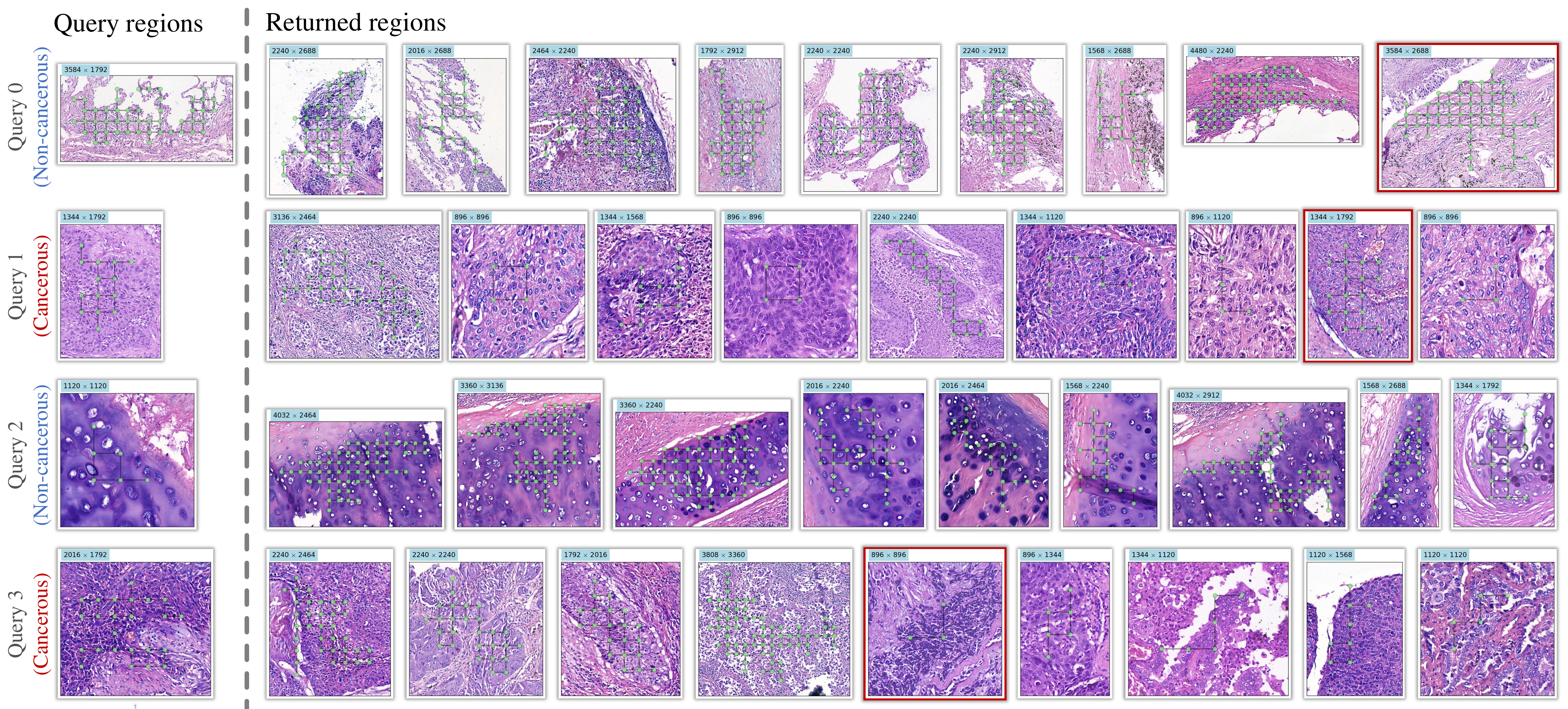}
	\caption{Visualization of the retrieval performance of the proposed method, where the first column provides the 4 query regions focused in Fig. \ref{F.tsne}, the top-returned regions from the retrieval are ranked on the right, the irrelevant return regions (has different labels with the query graph) are framed in red and the pixel size of the regions are located on the leftop of the images.} \label{F.visualization}
\end{figure*}

To study the allocation of graph codes in high-dimensional space, we employed the t-SNE \cite{maaten2008visualizing} tool to reduce the dimension of the outputs GCN-Hash model ($\mathbf{Y}$) and project the codes in 2-dimensional space. The visualization of the retrieval database is visualized in Fig. \ref{F.tsne}, where a dot represents a graph, the body color of the dot indicates the ratio of tumor occupation in the graph referring to the color bar on the right of the figure and the size of the dot is positive correction with the scale of graphs. Obviously, the \textit{Cancerous Graphs} and \textit{Cancer-free Graphs} are spread to the different sides of the space. The zoom-in regions on the corners of the figure focus on four retrieval instances, where the query graph is symbolized by hexagon with red boundary, the returned graphs are plotted with circles. Correspondingly, the structures of the graphs are presented. It is distinct that the query graphs are located nearby the database graphs with correct labels. Therefore, the relevant regions to the query region can be effectively retrieved by our model.

To further validate the qualitative performance of the proposed retrieval framework, we drew the graph structures on the retrieved regions. The joint visualization of WSI regions and graphs is provided in Fig.~\ref{F.visualization}. It shows that the relevant regions in various shape and size for the query region are returned from the WSIs. Especially for the query instance $2$ (the 3rd row in Fig.~\ref{F.visualization}), relevant regions containing 15 to 59 patches are successfully retrieved for the query region within 5 nodes (i.e. covers 5 patches). The results indicate the proposed model has learned the structural patterns of tissue regions. Therefore, it is qualified to the retrieval requirement in size and shape variations. 

\subsection{Discussion}
The appearances of query instances 0 and 2 in Fig.~\ref{F.visualization} share the same label (Cancer-free) in both the training of CNN and GCN models. While, the regions retrieved by our model also remain the corresponding appearance to each query region, i.e. the two different appearances are not confused in the retrieval model although the two appearances are told the same in the training. Supposing a classification or segmentation model trained using the same data, the predictions for the two tissue appearances are indistinguishable as they are labeled as the same thing (Cancer-free) in the training. Therefore, CBHIR is more promising in the aspect of providing assisted information than the image classification/segmentation methods to pathologists.

The encoding of regions in the proposed framework can be divided into 3 separate stages: feature extraction, graph construction and graph encoding. One of the future works will focus on combining the three stages into an integral model that can be trained end-to-end and can simultaneously predict the graph structures and the graph codes for a WSI. 

CBHIR is a passive auxiliary diagnosis process where the pathologist needs to find an interested region from the WSI and then call the CBHIR application. The diagnosis mode of auxiliary could be further improved by enhancing the autonomy of the framework. Therefore, another future work will focus on diagnostically relevant region recommendation algorithm based on the present CBHIR models for automatic diagnosis suggestion throughout the diagnosis procedure on digital pathology platforms.

\section{Conclusions}\label{S.conclusion}
In this paper, we proposed a novel histopathological image retrieval framework for large-scale WSI-database based on graph convolutional networks and the hashing technique. The instances in the database are defined based on graphs and are converted into binary codes by the designed GCN-Hash model. The experimental results have demonstrated that the GCN-model is scalable to size and shape variations of query regions and can effectively retrieve relevant regions that contain similar content and structure of tissue. It allows pathologists to create query regions by free-curves on the digital pathology platform. Benefiting from hashing structure, the retrieval process is completed based on hamming distance, which is very time-saving. The proposed model achieves the state-of-the-art retrieval performances on two public datasets involving breast and lung cancers when compared to the present frameworks for content-based histopathology image retrieval. One future work will peruse an integral model for graph construction and indexing of whole slide images. Another future work will focus on developing diagnostically relevant region recommendation algorithm to further improving the automation of the auxiliary diagnosis based on histopathological WSIs.

\section*{Acknowledgements}
This work was supported by the National Natural Science Foundation of China (grant number 61901018, 61771031, 61906058, and 61471016), the China Postdoctoral Science Foundation (grant number 2019M650446) and Tianjin Science and Technology Major Project (grant number 18ZXZNSY00260).

{\small
\bibliographystyle{ieee}
\bibliography{refs}
}

\end{document}